\documentclass[11pt]{article}

\usepackage[final]{acl}

\usepackage{times}
\usepackage{latexsym}

\usepackage[T1]{fontenc}

\usepackage[utf8]{inputenc}

\usepackage{microtype}

\usepackage{inconsolata}

\usepackage{graphicx}

\usepackage{CJKutf8}
\usepackage{newtxmath}
\usepackage{amsmath}
\usepackage{makecell}
\usepackage{array, multirow}
\usepackage[table, x11names, dvipsnames, xcdraw]{xcolor}
\usepackage[normalem]{ulem}
\useunder{\uline}{\ul}{}
\usepackage{booktabs}
\usepackage{nicematrix}
\usepackage{color, colortbl}
\usepackage{hhline}
\usepackage{subfigure}     

\definecolor{Gray}{gray}{0.9}

\usepackage{enumitem}
\usepackage[super]{nth}
\usepackage{amsfonts}

\usepackage{pmboxdraw}

\usepackage{appendix}

\usepackage{cleveref}
\usepackage{hyperref}

\title{\texttt{SCRIPT}: A Subcharacter Compositional Representation Injection Module\\ for Korean Pre-Trained Language Models}

\author{
  \textbf{SungHo Kim}\textsuperscript{1},
  \textbf{Juhyeong Park}\textsuperscript{1},
  \textbf{Eda Atalay}\textsuperscript{1},
  \textbf{SangKeun Lee}\textsuperscript{1,2}
\\
  \textsuperscript{1}Department of Artificial Intelligence, Korea University, Seoul, South Korea \\
  \textsuperscript{2}Department of Computer Science and Engineering, Korea University, Seoul, South Korea \\
\\
  \texttt{\{sungho3268, johnida, edaatalay, yalphy\}@korea.ac.kr}
}

\begin{document}

\begin{CJK}{UTF8}{mj}

\maketitle
\begin{abstract}
Korean is a morphologically rich language with a featural writing system in which each character is systematically composed of subcharacter units known as Jamo. 
These subcharacters not only determine the visual structure of Korean but also encode frequent and linguistically meaningful morphophonological processes. 
However, most current Korean language models (LMs) are based on subword tokenization schemes, which are not explicitly designed to capture the internal compositional structure of characters. 
To address this limitation, we propose \textbf{\texttt{SCRIPT}}, a model-agnostic module that injects subcharacter compositional knowledge into Korean PLMs.
\textbf{\texttt{SCRIPT}} allows to enhance subword embeddings with structural granularity, without requiring architectural changes or additional pre-training.
As a result, \textbf{\texttt{SCRIPT}} enhances all baselines across various Korean natural language understanding (NLU) and generation (NLG) tasks. Moreover, beyond performance gains, detailed linguistic analyses show that \textbf{\texttt{SCRIPT}} reshapes the embedding space in a way that better captures grammatical regularities and semantically cohesive variations. Our code is available at \href{https://github.com/SungHo3268/SCRIPT}{https://github.com/SungHo3268/SCRIPT}.

\end{abstract}

\section{Introduction}
In human writing systems, the grapheme, the smallest unit of written language that encodes linguistic information, plays a crucial role in shaping how meaning is represented and processed~\cite{coulmas2003writing, sampson2015writing, daniels1996world}. In many alphabetic systems, such as English, graphemes typically correspond to atomic letters (e.g., a, b, c), and words are formed through linear combination (e.g., “cat” consists of c, a, and t).
However, not all alphabetic systems operate in such a linear manner, nor do they necessarily treat characters as minimal units of written composition.

Korean, in particular, employs a unique featural writing system, \textit{Hangul}, in which each character is a structured composition of smaller subcharacter units known as Jamo. As illustrated in Figure~\ref{fig:Korean_Example.a}, each character consists of three \emph{Jamo} units: Choseong (initial consonant), Jungseong (vowel), and Jongseong (final consonant), following fixed spatial arrangements and a strict compositional order.
These principles were explicitly defined in \textit{Hunminjeongeum}\footnote{\textit{Hunminjeongeum} explains the letter design and well-formed combinations; modern encoding schemes and keyboard input sequences arise from contemporary standards.}~\cite{Hunminjeongeum_Guide, Hunminjeongeum_Haerye}, the historical document detailing the invention principles of \textit{Hangul}, including the design and combination rules for the subcharacters.

\definecolor{my_blue}{HTML}{0070C0}
\begin{figure*}[t!]
  \centering
  \subfigure[]{
    \includegraphics[width=0.22\linewidth]{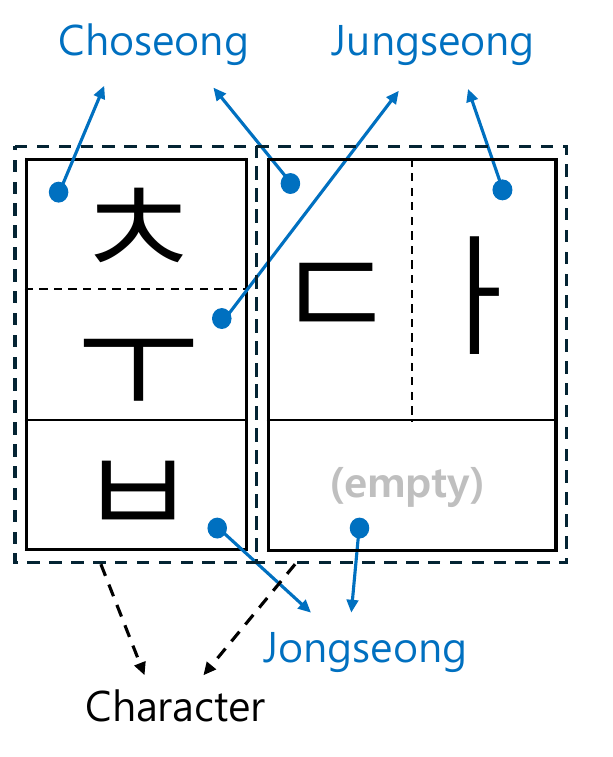}
    \label{fig:Korean_Example.a}
  }
  \subfigure[]{
    \includegraphics[width=0.73\linewidth]{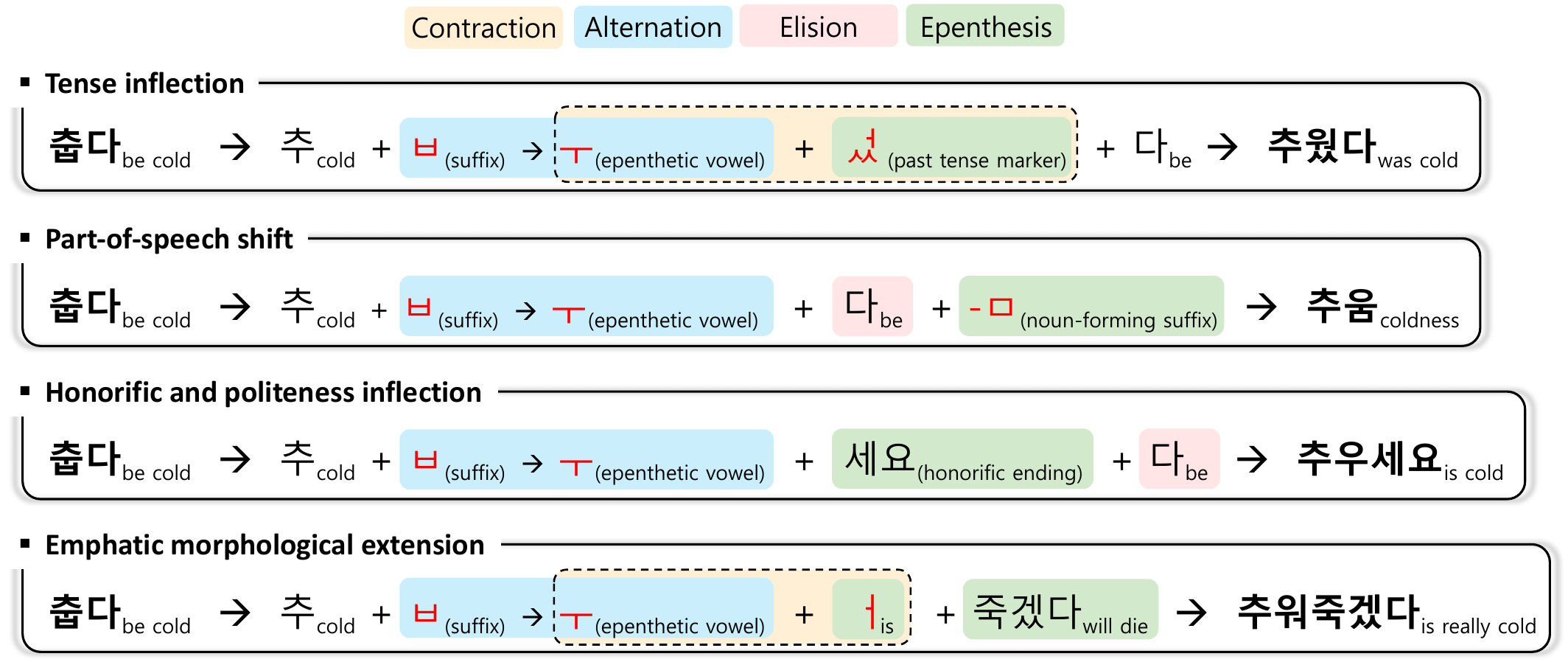}
    \label{fig:Korean_Example.b}
  }
  \caption{(a) Examples of the components of \textit{Hangul}. This figure illustrates two characters, such as `춥$_\text{cold}$' and `다$_\text{ending suffix}$', with each subcharacter highlighted in \textcolor{my_blue}{blue}. (b) Examples of linguistic phenomena arising from the inflection of predicate `춥다$_\text{be cold}$' at the subcharacter-level, with the transformed subcharacters highlighted in \textcolor{red}{red}.}
  \label{fig:Korean_Example}
\end{figure*}

Crucially, this compositional structure is not merely orthographic. As a morphologically rich and agglutinative language, Korean exhibits extensive morphophonological alternations across morpheme boundaries~\cite{lee2003modeling, matteson-etal-2018-rich, MorphoPhonologicalProcessesinKorean}. Predicate inflection, for example, often triggers systematic subcharacter-level alternations, such as the addition of the final consonant `ㅆ' to mark past tense or `-ㅁ' for nominalization, as shown in Figure~\ref{fig:Korean_Example.b}. Additionally, phonological assimilation between adjacent syllables frequently alters subcharacters to facilitate natural pronunciation~\cite{sohn2001korean, shin2012sounds}. These phenomena highlight that subcharacter-level features in Korean are tightly linked to grammatical, semantic, and morphophonological functions~\cite{lee2001korean}.

Despite this linguistic reality, most contemporary Korean PLMs, including advanced off-the-shelf LLMs~\cite{yoo2024hyperclovaxtechnicalreport, exaone35technicalreport}, rely almost exclusively on subword-based tokenization. While subword modeling effectively captures lexical semantics from large corpora, it struggles to reflect \textit{Hangul}’s compositional structure, limiting sensitivity to fine-grained morphosyntactic variations~\cite{albright2009predicting, kim-etal-2025-polishing}. In contrast, a few subcharacter-based LMs~\cite{moon-okazaki-2020-jamo, cognetta-etal-2023-parameter, kim-etal-2024-kombo} show strong robustness to such variations but often underperform on downstream tasks due to weaker semantic representations and increased computational cost.

To leverage complementary strengths, we propose \textbf{\texttt{SCRIPT}}, a lightweight, plug-and-play module that injects subcharacter-level structural knowledge directly into existing subword-based PLMs.
\textbf{\texttt{SCRIPT}} attaches to the embedding layer of a PLM and operates through a dual-channel strategy. 
It compresses subcharacter sequences into structure-aware subword representations grounded in \textit{Hangul}’s compositional principles, and then fuses them with the PLM’s original subword embeddings. 
This design enables the model to capture fine-grained subcharacter compositionality while preserving the rich semantic information learned from large-scale corpora, without modifying the PLM architecture or requiring additional pre-training.
The main contributions of this work are summarized as follows:
\begin{itemize}  [leftmargin=15px, 
]
    \item We empirically show that most Korean morphological variations occur at the subcharacter-level, motivating subcharacter-aware modeling.
    \item We introduce \textbf{\texttt{SCRIPT}}, a model-agnostic module that injects structure-aware subcharacter compositional representations into existing PLMs via embedding-level integration.
    \item We show that \textbf{\texttt{SCRIPT}} improves performance across a wide range of Korean NLU and NLG benchmarks, while effectively capturing key morphosyntactic phenomena.
\end{itemize}

\begin{figure*}[t!]
  \centering
  \setlength{\abovecaptionskip}{-3pt} 
  \includegraphics[width=1.0\linewidth]{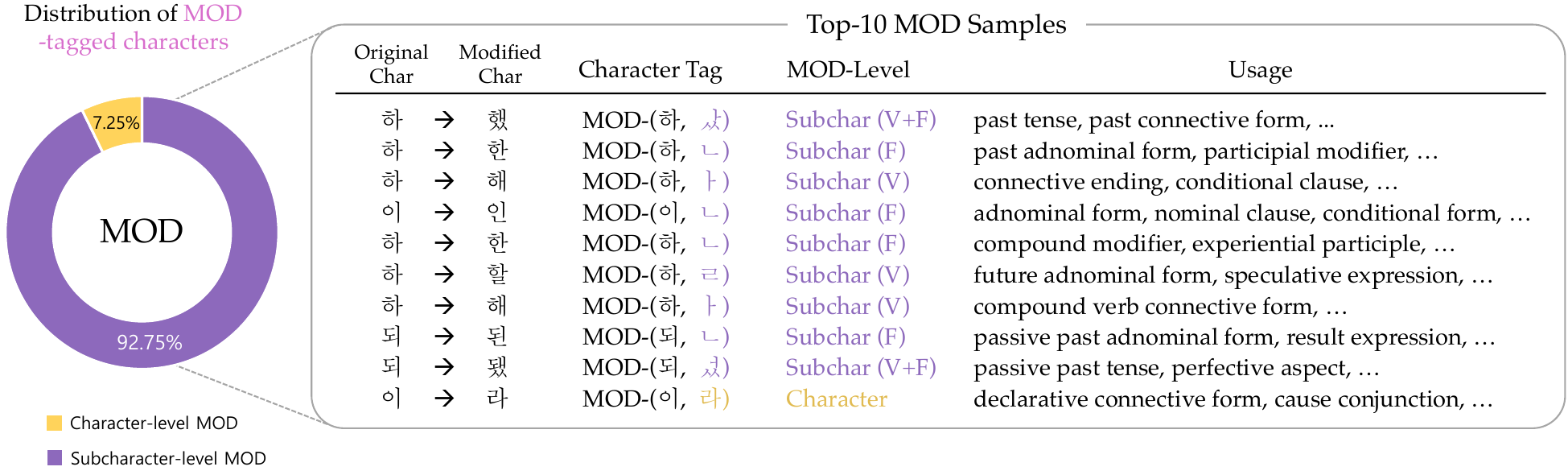}
  \label{f}
  \caption{Morphological modifications in a large-scale Korean POS-tagged corpus: the left panel distinguishes subcharacter- and character-level \texttt{MOD} cases, and the right lists the ten most frequent \texttt{MOD} types with raw target characters and corresponding factors.}
  \label{fig:morphophonological_analysis}
\end{figure*}

\section{Motivation}
\label{sec:motivation}
In this section, we present our empirical observations on the pervasiveness of diverse morphophonological changes at the subcharacter-level in real Korean usage, highlighting the importance of modeling the subcharacter structure of \textit{Hangul}.

\subsection{Setup}
\label{sec:motivation.setup}
Specifically, we conduct a large-scale corpus-based analysis quantifying their frequency. We used the Korean Part-of-Speech Tagged Corpus\footnote{Part-of-Speech Tagged Corpus (v1.1) from \href{https://kli.korean.go.kr/}{ModuCorpus}, provided by the National Institute of Korean Language (2020)}, which contains 3M words annotated with morpheme-level and POS information. This corpus provides a solid basis for estimating the frequency of subcharacter-level alternations in real usage.

To systematically capture these alternations, we adopted a simple annotation scheme, following \citet{matteson-etal-2018-rich}, which marks how each character relates to its base form (lemma). Each character is assigned to one of three categories:
\begin{itemize} [leftmargin=15px, itemsep=0em]
    \item \texttt{KEEP}: unchanged with respect to the base form.
    \item \texttt{MOD}: modified from the base form.
    \item \texttt{NOOP}: omitted in the base form.
\end{itemize}
For example, in `했다$_\text{did}$', whose base form is `하다$_\text{do}$', the segment `했$_\text{did}$' is labeled as \texttt{MOD} because it reflects a tense change arising from the combination of the verb stem `하' and the past tense marker `-었-', which undergoes phonological contraction (`하 + 었 → 했'). In contrast, `다' is labeled as \texttt{KEEP} since it remains unchanged.

In this paper, we focus our analysis on characters tagged as \texttt{MOD}, as they directly encode morphological alternations. 
Each \texttt{MOD} character is further classified by its level of granularity~\footnote{Detailed tagging procedures are provided in Appendix~\ref{app:inflection_freq}}:
\begin{itemize} [leftmargin=15px, itemsep=0.2em]
    \item \textbf{Subcharacter-level \texttt{MOD}}: when only part of a character is altered. (e.g., the change from `하' to `한', adding the final consonant `ㄴ').
    \item \textbf{Character-level \texttt{MOD}}: when the entire character is changed into a different character. (e.g., `이' is replaced with `라').
\end{itemize}

\subsection{Observation}
\label{sec:motivation.observation}
As shown in Figure~\ref{fig:morphophonological_analysis}, the overwhelming majority of characters tagged as \texttt{MOD} (92.75\%) involved subcharacter-level modifications, while only a small fraction (7.25\%) represented character-level changes. 
These findings underscore the importance of modeling subcharacter-level alternations for a deeper and more comprehensive understanding of Korean, especially in adapting to diverse usage and morphological variation. This aligns with prior work~\cite{kim-etal-2024-kombo, lee-etal-2025-jamo}, which observed that jamo-based language modeling demonstrates robustness in handling character-level conjugation changes and exhibits strong performance on noisy, real-world data such as offensive content.
Motivated by this, we aim to explicitly encode subcharacter compositional knowledge in a principled manner, thereby incorporating this linguistic information into language models that have previously overlooked it.

\begin{figure*}[t!]
  \setlength{\abovecaptionskip}{-1pt} 
  \centering
  \subfigure[]{
    \includegraphics[width=.28\linewidth]{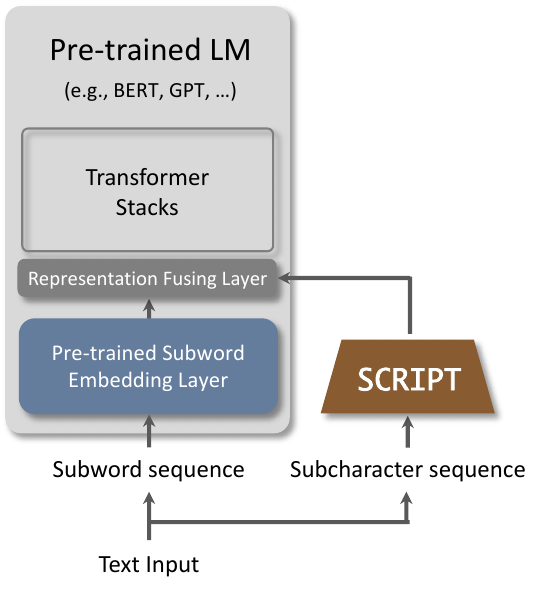}
    \label{fig:script_overall}
  }
  \subfigure[]{
    \includegraphics[width=.66\linewidth]{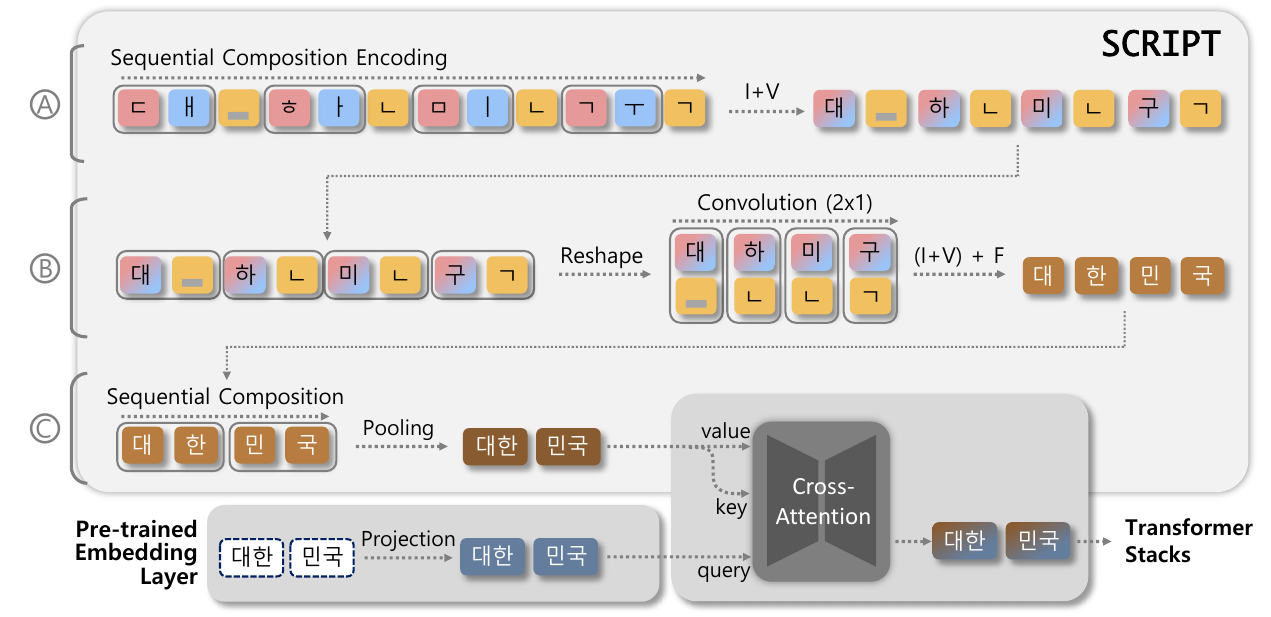}
    \label{fig:script_details}
  }
  \caption{(a) Overall illustration of the PLM enhanced with \textbf{\texttt{SCRIPT}}. (b) Detailed architectural example of the \textbf{\texttt{SCRIPT}}, starting from the word `대한민국$_\text{South Korea}$', which is tokenized into 12 Jamo units: \,I\,: [ㄷ, ㅎ, ㅁ, ㄱ],\; V: [ㅐ, ㅏ, ㅣ, ㅜ\;],\; F: [\;{\tiny \color{gray} \pmboxdrawuni{2583}}\;, ㄴ, ㄴ, ㄱ]. Each sub-process (A-C) represents a successive fusion step: (A) fusion of Choseong and Jungseong (\S\ref{sec:cho+jung}), (B) addition of Jongseong (\S\ref{sec:add_jong}), and (C) character-to-subword (\S\ref{sec:char_to_subword}).}
  \label{fig:script_architecture}
\end{figure*}

\section{Methodology}
\label{sec:methodology}
Based on our observation (\S\ref{sec:motivation}), we propose \textbf{\texttt{SCRIPT}} (\textbf{S}ubcharacter \textbf{C}ompositional \textbf{R}epresentation \textbf{I}njection Module for Korean \textbf{P}re-\textbf{T}rained Language Model), 
a module that enhances PLM's embeddings with subcharacter compositional knowledge.
In this section, we instance Jamo as the subcharacter unit in \textbf{\texttt{SCRIPT}} and apply it to subword-based PLM, aligning with standard practices in modern Korean PLMs. We also provide extensions for alternative subcharacter units, such as BTS units (Appendix~\ref{app:bts_units}).

\subsection{Overall Framework}
\label{sec:overall_model_architecture}

\textbf{\texttt{SCRIPT}} is attached to PLMs at the embedding layer, as illustrated in Figure~\ref{fig:script_overall}. Given a Korean text, the model employs two parallel tokenization paths: (1) a subword tokenizer that produces the PLM's original subword sequence, and (2) a subcharacter tokenizer that generates fine-grained input for \textbf{\texttt{SCRIPT}}. The subword sequence is projected through the PLM’s original embedding layer, while \textbf{\texttt{SCRIPT}} constructs an alternative subword-level representation by compressing the subcharacter sequence (\S\ref{sec:script_architecture}). 
These two subword representations are then integrated into a unified subword representation (\S\ref{sec:fusion_subword}).
This dual-channel strategy allows us to leverage the strengths of both approaches.
The full algorithm for synthesizing subword representations with \textbf{\texttt{SCRIPT}} is provided in Table~\ref{tab:script_algo}.

\subsection{\textbf{\texttt{SCRIPT}}}
\label{sec:script_architecture}

Figure~\ref{fig:script_details} illustrates the detailed architecture of \textbf{\texttt{SCRIPT}}, which compresses subcharacter representations into subword representations in two stages.

\subsubsection{Stage 1: Subcharacter-to-Character}
\label{sec:subchar_to_char}
The first stage of deriving subword representations from subcharacter representations is to compress subcharacter representations into character representations. Inspired by ~\citet{kim-etal-2024-kombo}, to effectively model the compositional structure of \textit{Hangul}, we explicitly incorporate three fundamental compositional principles into our methodology~\cite{Hunminjeongeum_Haerye, yeon2013korean, unicode12}:

\begin{enumerate} [leftmargin=20px, itemsep=0.2em]  \label{sec:fundamental_principles}
    \item \textit{\textbf{Composition}}: A character is composed of up to three Jamo: Choseong and Jungseong are essential components, whereas Jongseong is not mandatory.~\footnote{A detailed explanation is provided in Appendix~\ref{app:hangul_character_explanation}}
    \item \textit{\textbf{Spatial arrangement}}: Within a syllable block, Choseong is placed either above or to the left of Jungseong, while Jongseong, if present, is always positioned beneath them.
    \item \textit{\textbf{Sequential order}}: Jamo consistently follow a prescribed order: Choseong $\rightarrow$ Jungseong $\rightarrow$ Jongseong.
\end{enumerate}

\textbf{\texttt{SCRIPT}} adopts a hierarchical compression architecture grounded in the design principles of \textit{Hangul}, to better capture linguistic features of Korean. This step underpins the `structure-aware' subword embeddings in \textbf{\texttt{SCRIPT}}, as it explicitly encodes the subcharacter compositional knowledge.

\paragraph{Subcharacter Representation.}
Given an input text $\mathbf{s}$, each character is first decomposed into sequential three subcharacters, I (short for initial consonant, Choseong), V (short for vowel, Jungseong), and F (short for final consonant, Jongseong), following \textit{Principles 1}. 
If a character lacks a final consonant, a special empty token ({\tiny \color{gray} \pmboxdrawuni{2583}}) is inserted in its place. 
The resulting subcharacter embeddings are denoted as $\mathbf{e} \in \mathbb{R}^{N \times D}$, where $N$ is the number of subcharacter tokens and $D$ is the embedding dimension.
Furthermore, to clarify the ordered arrangement of I, V, and F, we denote the sequential subcharacter representation $e_i\in\mathbf{e}$ as $e_{\text{I}, k}$, $e_{\text{V}, k}$, and $e_{\text{F}, k}$ for each integer $k$ in [1, N/3]:
\begin{equation} \label{eq:3}
  e_i=
    \begin{cases}
    e_{\text{I}, k} & \mbox{if } i = 3k-2  \\
    e_{\text{V}, k} & \mbox{if } i = 3k-1  \\
    e_{\text{F}, k} & \mbox{if } i = 3k  
    \end{cases}
\end{equation}

\paragraph{Fusion of Choseong and Jungseong.}
\label{sec:cho+jung}
To accurately reflect the sequential order (\textit{Principle 3}), we follow the fixed composition sequence (I $\rightarrow$ V $\rightarrow$ F). The entire subcharacter sequence is first encoded using a GRU-based sequential composition layer. Following this order, the I and V components are merged via element-wise summation to obtain a combined representation, $\mathbf{h}_{\text{I+V}} \in \mathbb{R}^{\frac{N}{3} \times D}$.

\paragraph{Addition of Jongseong.}
\label{sec:add_jong}
After forming the intermediate representation of I and V, we incorporate the V to complete the character representation. Reflecting the visual structure of \textit{Hangul}, where F is always positioned below I and V (\textit{Principle 2}), we model this arrangement by vertically concatenating the F representation, $\textbf{h}_\text{F}=\{h_{\text{F},k}\}$, with $\mathbf{h}_\text{I+V}$. This composition is formally expressed as follows:
\begin{equation} \label{eq:6}
    \mathbf{h}_\text{R}=
    \begin{bmatrix}
        \mathbf{h}_\text{I+V} \\
        \mathbf{h}_\text{F}
    \end{bmatrix}
    \quad \in\mathbb{R}^{2 \times \frac{N}{3} \times D}
\end{equation}
To merge these vertically aligned subcharacters into characters, we apply a convolutional layer capturing the relative positional information. We then finalize the character representations by applying average pooling over $\textbf{h}_\text{R}$, yielding dense character representations, $\textbf{h}_\text{C} \in\mathbb{R}^{\frac{N}{3} \times D}$, grounded in the compositional principles of \textit{Hangul}.

\subsubsection{Stage 2: Character-to-Subword}
\label{sec:char_to_subword}
The second main stage of \textbf{\texttt{SCRIPT}} compresses character representations into subword representations, aligning their granularity with that of the original subwords used in PLMs. Our goal was to aggregate character representations within each subword to form a unified subword representation. However, directly averaging or summing these character representations often led to unstable training.
To mitigate this issue, and in line with \textit{Principle 3}, we apply the sequential composition layer once more to capture the compositional order of characters within each subword. Then, we apply a simple pooling operation, specifically, selecting the final character representation at each subword boundary, to obtain the subword representation:
\begin{equation} \label{eq:8}
    \mathbf{h}_\text{S} = \textsc{Pooling}(\textsc{GRU}(\mathbf{h}_\text{C}))
    \quad \in\mathbb{R}^{N' \times D}
\end{equation}
where $N'$ denotes the subword sequence length.

\subsection{Fusion of Two Subword Representations}
\label{sec:fusion_subword}
Despite these structured, dense subcharacter-level linguistic features, the resulting subword representations, compressed from subcharacters alone, lack semantic expressiveness, as they are not pre-trained on large-scale Korean corpora. To address this limitation, we fuse them with semantically richer subword embeddings obtained from the existing PLM. Specifically, we introduce a fusion mechanism that integrates two complementary representations: the synthesized subword representation from \textbf{\texttt{SCRIPT}}, denoted as $\mathbf{h_{S}}$, and the original pre-trained subword embedding, $\mathbf{e_{S}} \in \mathbb{R}^{N' \times D}$, projected into the same embedding space. A cross-attention layer is employed to combine these sources, yielding the final structure-aware subword representation $\mathbf{e_{F}} \in \mathbb{R}^{N' \times D}$, which is then used as input to the subsequent Transformer layers:
\begin{equation} \label{eq:11}
    \mathbf{e}_\text{F} = \textsc{CrossAttn}(\mathrm{Q}=\mathbf{e_{S}}, \mathrm{KV}=\mathbf{h_{S}})
\end{equation}

Through \textbf{\texttt{SCRIPT}}, we construct a fused subword representation that integrates the compositional knowledge of \textit{Hangul} with the semantic richness of pre-trained subword embeddings. This dual-channel encoding enhances the language model’s ability to capture Korean character structure while preserving subword-level semantic content. These fused representations are then fed into the PLM’s Transformer stacks, allowing downstream tasks to benefit from this linguistically enriched input.

\begin{table*}[t!]
  \centering
  \small
  \setlength{\tabcolsep}{5pt}         
  \renewcommand{\arraystretch}{1.}    
  \begin{tabular}{lccccccccc}
    \toprule
    \multirow{2.5}{*}{Model} & \multirow{2.5}{*}{KorNLI} & \multirow{2.5}{*}{KorSTS} & \multirow{2.5}{*}{NSMC} & \multirow{2.5}{*}{PAWS-X} & \multicolumn{5}{c}{KoBEST} \\
    \cmidrule(lr){6-10}
     & & & & & 
    \multicolumn{1}{c}{BoolQ} & \multicolumn{1}{c}{COPA} & \multicolumn{1}{c}{WiC} & \multicolumn{1}{c}{HellaSwag} & \multicolumn{1}{c}{SentiNeg} \\
    \midrule
    \multicolumn{1}{l}{KOMBO$_\text{base}$} & 75.97 & 77.28 & 88.34 & 73.40 & 61.40 & 61.00 & 68.91 & 63.80 & 79.07 \\
    \multicolumn{1}{l}{BERT$_\text{base}$} & 75.85 & 76.72 & \textbf{88.96} & 72.38 & 60.75 & 60.90 & 73.14 & 63.20 & 83.12 \\
    \multicolumn{1}{l}{BERT$_\text{base}$ + \textbf{\texttt{SCRIPT}} } & \textbf{76.49} & \textbf{77.68} & \textbf{88.96} & \textbf{73.68} & \textbf{62.32} & \textbf{61.30} & \textbf{74.30} & \textbf{64.40} & \textbf{83.38} \\
    \midrule
    \multicolumn{1}{l}{KoGPT2$_\text{base}$} & 72.24 & 73.82 & \textbf{88.90} & 76.33 & 67.22 & 68.90 & 67.07 & 69.10 & 88.50 \\
    \multicolumn{1}{l}{KoGPT2$_\text{base}$ + \textbf{\texttt{SCRIPT}}} & \textbf{72.47} & \textbf{74.27} & 88.80 & \textbf{76.61} & \textbf{68.28} & \textbf{70.90} & \textbf{68.18} & \textbf{72.40} & \textbf{89.47} \\
    \midrule
    \multicolumn{1}{l}{KoGPT3-1.2B} & 80.11 & 76.14 & 90.51 & 77.40 & 77.32 & \textbf{82.80} & 72.78 & 78.90 & 96.31 \\
    \multicolumn{1}{l}{KoGPT3-1.2B + \textbf{\texttt{SCRIPT}}} & \textbf{80.39} & \textbf{79.60} & \textbf{90.53} & \textbf{79.95} & \textbf{77.63} & \textbf{82.80} & \textbf{74.65} & \textbf{79.30} & \textbf{\textbf{96.48}} \\
    \midrule
    \multicolumn{1}{l}{EXAONE-2.4B} & 83.99 & 85.08 & 90.04 & 85.24 & 92.59 & \textbf{\textbf{93.30}} & 82.14 & 85.60 & 94.21 \\
    \multicolumn{1}{l}{EXAONE-2.4B + \textbf{\texttt{SCRIPT}}} & \textbf{85.77} & \textbf{85.27} & \textbf{90.89} & \textbf{85.90} & \textbf{93.30} & \textbf{93.30} & \textbf{82.46} & \textbf{86.00} & \textbf{94.96} \\
    \bottomrule
  \end{tabular}
  \caption{Performance on nine Korean NLU tasks. The evaluation metrics for each task are as follows: 
  KorSTS is evaluated using Spearman correlation $\times$100, while other tasks are evaluated based on accuracy (\%). The best results in each family of models are highlighted in \textbf{boldface}.}
  \label{tab:nlu_results}
\end{table*}

\begin{table*}[h!]
  \centering
  \small
  \setlength{\tabcolsep}{5pt}       
  \renewcommand{\arraystretch}{1.}    
  \begin{tabular}{lccccccc}
    \toprule
    \multirow{2.5}{*}{Model} & \multicolumn{7}{c}{KoCommonGen} \\
    \cmidrule(lr){2-8} & 
    \multicolumn{1}{c}{BLEU 3} & \multicolumn{1}{c}{BLEU 4} & \multicolumn{1}{c}{ROUGE-2} & \multicolumn{1}{c}{ROUGE-L} & \multicolumn{1}{c}{METEOR} & \multicolumn{1}{c}{mBERTScore} & \multicolumn{1}{c}{KoBERTScore} \\ 
    \midrule
    \multicolumn{1}{l}{KoGPT2$_\text{base}$} & 18.29 & 10.33 & 44.24 & 54.50 & 40.05 & 83.37 & 91.21 \\
    \multicolumn{1}{l}{KoGPT2$_\text{base}$ + \textbf{\texttt{SCRIPT}}} & \textbf{25.01} & \textbf{15.57} & \textbf{47.42} & \textbf{60.00}  & \textbf{42.53} & \textbf{84.67} & \textbf{91.46}\\
    \midrule
    \multicolumn{1}{l}{KoGPT3-1.2B} & 26.19 & 17.20 & 58.85 & 62.53 & 52.11 & 85.41 & 91.17\\
    \multicolumn{1}{l}{KoGPT3-1.2B + \textbf{\texttt{SCRIPT}}} & {\textbf{28.89}} & {\textbf{19.58}} & {\textbf{59.28}} & {\textbf{64.80}} & {\textbf{52.37}} & {\textbf{86.26}} & {\textbf{91.78}}\\
    \midrule
    \multicolumn{1}{l}{EXAONE-2.4B} & 40.11 & 28.41 & 62.25 & 64.84 & 54.84 & 87.65 & 93.12 \\
    \multicolumn{1}{l}{EXAONE-2.4B + \textbf{\texttt{SCRIPT}}} & \textbf{41.48} & \textbf{31.80} & \textbf{71.03} & \textbf{72.16} & \textbf{61.27} & \textbf{88.12} & \textbf{93.95}\\
    \bottomrule
  \end{tabular}
  \caption{Performance on KoCommonGen generative task. We use eight automatic evaluation metrics, including n-gram based measures like BLEU, ROUGE, and METEOR, and two BERT-based scores for semantic similarity. The best results in each family of models are highlighted in \textbf{boldface}.}
  \label{tab:kocommongen}
\end{table*}

\section{Experiments}
\label{sec:experiments}
In this section, we evaluate \textbf{\texttt{SCRIPT}} on a range of Korean NLU and NLG tasks across strong PLMs (\S\ref{sec:exp_results}). We further conduct ablation studies to analyze the contribution of \textit{Hangul}-specific structural knowledge and key design choices (\S\ref{sec:ablation_study}). 

Additionally, we show that this efficiency comes with minimal computational overhead, which remains comparable to standard subword-based models (see Appendix~\ref{app:computational_efficiency}).

\subsection{Experimental Settings}
\label{sec:exp_settings}
\paragraph{Baselines.}
We applied \textbf{\texttt{SCRIPT}} to four Korean subword-based PLMs (KoGPT2$_\text{base}$, KoGPT3-1.2B, EXAONE-3.5-2.4B-Instruct, BERT$_\text{base}$), and additionally compare with a state-of-the-art Jamo-based encoder model, KOMBO$_\text{base}$ \cite{kim-etal-2024-kombo}. 
Detailed specifications and implementational details are provided in Appendix~\ref{app:model_consideration},~\ref{app:experimental_settings}.

\paragraph{Tasks.}
We evaluate \textbf{\texttt{SCRIPT}}-enhanced models on nine Korean NLU tasks, including four standard benchmarks (KorNLI, KorSTS, NSMC, PAWS-X) and five KoBEST tasks designed to assess diverse linguistic and cognitive capabilities. To evaluate generative performance, we additionally consider KoCommonGen for commonsense reasoning, XL-Sum for summarization, and Korean GEC for grammatical error correction. Detailed dataset statistics and explanations are provided in Appendix~\ref{app:tasks}.

\subsection{Experimental Results}
\label{sec:exp_results}
\paragraph{Korean Standard NLU Tasks.}
\label{sec:korean_standard_nlu_tasks}
As shown in Table~\ref{tab:nlu_results}, \textbf{\texttt{SCRIPT}} improves performance across all baselines, yielding average gains of up to 1.6\%p. Compared to the Jamo-based baseline KOMBO$_\text{base}$, our BERT$_\text{base}$+\textbf{\texttt{SCRIPT}} model achieves superior performance despite using the same underlying architecture and a comparable model size. Unlike KOMBO$_\text{base}$, which directly processes raw subcharacters and relies on costly full pre-training, \textbf{\texttt{SCRIPT}} is applied as a plug-in module during only fine-tuning, enabling efficient incorporation of \textit{Hangul} structure. 
Notably, \textbf{\texttt{SCRIPT}} is applicable to both encoder and decoder architectures, overall improving performance across model types.

\paragraph{Korean Advanced NLU Tasks.}
\label{sec:korean_advanced_nlu_tasks}
\textbf{\texttt{SCRIPT}} also outperforms baselines on knowledge-intensive tasks in KoBEST, including reading comprehension (KB-WiC) and commonsense reasoning (KB-HellaSwag). It proves more effective than KOMBO$_\text{base}$, which lags on complex tasks due to its reliance on subcharacter-only inputs. By integrating subword and subcharacter information, \textbf{\texttt{SCRIPT}} offers robust and architecture-agnostic enhancements across task complexities.

\paragraph{Korean Generation Tasks.}
\label{sec:korean_nlg_tasks}
Our method also delivers consistent gains on Korean generative tasks. As shown in Table~\ref{tab:kocommongen}, on KoCommonGen, a task that involves transforming and combining given morphemes to generate plausible sentences, \textbf{\texttt{SCRIPT}} improves across all seven generative metrics, with gains (an average of 1.4-3.5\%p) depending on model size. The improvements are particularly pronounced in n-gram metrics such as BLEU, METEOR, and ROUGE, indicating enhanced modeling of local compositional patterns in morphologically rich Korean.
This trend extends to other generation tasks as shown in Appendix~\ref{app:extra_gen_tasks}. Notably, on the Korean GEC task Kor-Learner, \textbf{\texttt{SCRIPT}} exceeds the best-performing baseline by over 3.2\%p on average. According to \citet{yoon-etal-2023-towards}, Kor-Learner includes a high concentration of errors involving particles, endings, and conjugations compared to Kor-Native. 

One possible explanation for the relatively larger gains on generative tasks, compared to NLU tasks, is that Hangul’s sequential compositional structure (Choseong $\rightarrow$ Jungseong $\rightarrow$ Jongseong) aligns naturally with token-by-token decoding, allowing subcharacter information to more directly influence generation decisions. We consider this a promising direction for further analysis, as a deeper understanding of this phenomenon requires further investigation.

\begin{table*}[t!]
  \small
  \centering
  \resizebox{\textwidth}{!}{\begin{tabular}{ccc|cccccc}
    \toprule
    \multicolumn{2}{c}{\textbf{\texttt{SCRIPT}}} & \multirow{2.5}{*}{Fusion w/ PLM} & \multicolumn{5}{c}{KoBEST} & \multirow{2.5}{*}{Avg.} \\
    \cmidrule(lr){1-2} \cmidrule(lr){4-8}
    Initial Token Unit & Compression &  & \multicolumn{1}{c}{BoolQ} & \multicolumn{1}{c}{COPA} & \multicolumn{1}{c}{WiC} & \multicolumn{1}{c}{HellaSwag} & \multicolumn{1}{c}{SentiNeg} \\
    \midrule

    \multicolumn{1}{c}{Jamo} & Principles & CrossAttention & {68.28} & \textbf{{70.90}} & \textbf{{68.18}} & \textbf{{72.40}} & \textbf{{89.47}} & \textbf{{73.85}} \\
    \midrule

    \cellcolor[HTML]{EAECFF}{Stroke} & Principles & CrossAttention & 68.35 & {65.20} & 67.81 & 71.00 & {89.46} & {72.36} \\
    \cellcolor[HTML]{EAECFF}{Cji} & Principles & CrossAttention & \textbf{{68.82}} & {65.20} & {67.98} & 71.40 & 88.01 & 72.28 \\
    \cellcolor[HTML]{EAECFF}{BTS} & Principles & CrossAttention & 67.99 & 64.70 & 66.67 & {71.50} & 88.97 & 71.97 \\
    \midrule

    \cellcolor[HTML]{EAECFF}{Character} & Principles & CrossAttention & 66.00 & 59.30 & 63.37 & 69.40 & 88.40 & 69.29 \\
    \cellcolor[HTML]{EAECFF}{Subword} & Principles & CrossAttention & 59.19 & 54.50 & 67.62 & 69.70 & 88.57 & 67.92 \\
    \cellcolor[HTML]{EAECFF}{Word} & Principles & CrossAttention & 66.48 & 61.10 & 62.82 & 72.00 & 88.71 & 70.22 \\
    \midrule
    
    \multicolumn{1}{c}{Jamo} & \cellcolor[HTML]{EAECFF}{Linear} & CrossAttention & 67.04 & 57.70 & 64.35 & 69.30 & 88.11 & 69.30 \\
    \multicolumn{1}{c}{Jamo} & \cellcolor[HTML]{EAECFF}{Attention} & CrossAttention & {68.14} & {64.70} & {67.28} & {71.60} & {88.32} & {72.01} \\
    \midrule
        
    \multicolumn{1}{c}{Jamo} & Principles & \cellcolor[HTML]{EAECFF}{Summation} & {68.27} & {65.50} & {67.98} & {70.60} & {89.07} & {72.28} \\
    \multicolumn{1}{c}{Jamo} & Principles & \cellcolor[HTML]{EAECFF}{Concatenation} & 66.55 & 61.10 & 66.61 & 69.90 & 89.06 & 70.64 \\
    \bottomrule
  \end{tabular}}
  \caption{Ablation results for various architecture of \textbf{\texttt{SCRIPT}} applied to KoGPT2$_\text{base}$. The first row presents the best-performing variant. Cells corresponding to ablated components are highlighted in \colorbox[HTML]{EAECFF}{\strut light blue}. The global-best results are highlighted in \textbf{boldface}.}
  \label{tab:ablation_main}
\end{table*}

\subsection{Ablation Study for \textbf{\texttt{SCRIPT}} Architecture}
\label{sec:ablation_study}
Table~\ref{tab:ablation_main} presents an ablation study examining the core design choices of \textbf{\texttt{SCRIPT}}.

\paragraph{Alternative Tokenization Methods for \textbf{\texttt{SCRIPT}}.}
Across different granularities, Jamo-based \textbf{\texttt{SCRIPT}} achieves the best overall performance. In contrast, using excessively fine-grained units such as BTS leads to a slight performance drop, suggesting limitations in compressing overly fine-grained representations into higher-level coarse units.
Furthermore, when we extended the comparison to larger units beyond the subcharacter level, including character, subword, and word units, performance degraded substantially. This finding suggests that our proposed method is specifically designed to preserve the compositional structure of subcharacters and is therefore less suited to larger linguistic units. Notably, using subword units, also employed in the base PLM, resulted in the largest performance drop. This result indicates that the observed gains are not simply attributable to increased parameter count or additional fusion capacity, but are instead driven by Jamo-level structural information.

\paragraph{Compression Method of Subcharacters in \textbf{\texttt{SCRIPT}}.}
Replacing the proposed composition-principled compression with generic pooling methods (Attention~\cite{funnel-transformer2020} or Linear~\cite{nawrot-etal-2022-hierarchical}) leads to a 1.8–4.6\%p performance drop, underscoring the importance of preserving the hierarchical compositional structure of \textit{Hangul} during subcharacter aggregation.

\paragraph{Integration Method of Subword Representations.}
We further compare integration strategies between subcharacter and subword representations. CrossAttention~\cite{transformer2017} yields the strongest results, outperforming Summation and Concatenation, suggesting that dynamic alignment with PLM representations is key to integrating two heterogeneous subword representations effectively.

Overall, these results demonstrate that \textbf{\texttt{SCRIPT}}’s gains arise from explicitly encoding \textit{Hangul}’s compositional structure and aligning it with pre-trained representations, rather than from any single architectural choice.

\begin{table*}[t!]
    \centering
    \small
    \setlength{\tabcolsep}{5pt}         
    \renewcommand{\arraystretch}{1.1}    
    \begin{tabular}{lccccccl}
        \toprule
        \multicolumn{1}{c}{\multirow{2.5}{*}{Model}} & \multirow{2.5}{*}{\begin{tabular}[c]{@{}c@{}}PLM\\Tokenization \end{tabular}} & \multicolumn{5}{c}{KoBEST} & \multicolumn{1}{c}{\multirow{2.5}{*}{Avg.}} \\ 
        \cmidrule{3-7}
        \multicolumn{1}{c}{} &  & BoolQ & COPA & WiC & HellaSwag & SentiNeg &  \\ 
        \midrule
        BERT$_\text{base}$ & \multirow{2}{*}{\begin{tabular}[c]{@{}c@{}}Word\end{tabular}} & 60.04 & {\ul 57.60} & 62.70 & 55.00 & {\ul 52.39} & 57.55 \\
        BERT$_\text{base}$ +  \textbf{\texttt{SCRIPT}}$_\text{Jamo}$ &  & {\ul 60.47} & {\ul 57.60} & {\ul 64.76} & {\ul 57.20} & {\ul 52.39} & {\ul 58.48}({\color{ForestGreen} $\text{▲}$} 0.94) \\ 
        \midrule
        BERT$_\text{base}$ & \multirow{2}{*}{\begin{tabular}[c]{@{}c@{}}Morpheme\end{tabular}} & 63.75 & 58.50 & 71.75 & 61.80 & 78.59 & 66.88 \\
        BERT$_\text{base}$ +  \textbf{\texttt{SCRIPT}}$_\text{Jamo}$ &  & {\ul 65.03} & {\ul 60.00} & {\ul \textbf{72.06}} & {\ul 62.20} & {\ul 80.35} & {\ul 67.93}({\color{ForestGreen} $\text{▲}$} 1.05) \\ 
        \midrule
        BERT$_\text{base}$ & \multirow{2}{*}{\begin{tabular}[c]{@{}c@{}}Subword\end{tabular}} & 67.22 & 67.10 & 68.90 & 69.10 & 88.50 & 72.16 \\
        BERT$_\text{base}$ +  \textbf{\texttt{SCRIPT}}$_\text{Jamo}$ &  & \textbf{{\ul 68.28}} & \textbf{{\ul 68.20}} & {\ul 70.90} & \textbf{{\ul 72.40}} & \textbf{{\ul 89.47}} & \textbf{{\ul 73.85}}({\color{ForestGreen} $\text{▲}$} \textbf{1.69}) \\ 
        \midrule
        BERT$_\text{base}$ & \multirow{2}{*}{\begin{tabular}[c]{@{}c@{}}Character\end{tabular}} & {\ul 62.89} & {\ul 61.00} & {\ul 71.35} & 48.60 & {\ul 78.84} & {\ul 64.54} \\
        BERT$_\text{base}$ +  \textbf{\texttt{SCRIPT}}$_\text{Jamo}$ &  & 61.04 & 59.30 & {\ul 71.35} & {\ul 49.40} & 78.34 & 63.89({\color{Red} $\text{▼}$} 0.65) \\ 
        \bottomrule
    \end{tabular}
    \caption{Comparison of the effectiveness of compositional knowledge integration into PLMs across different tokenization methods. The global-best results are highlighted in \textbf{boldface} and local-best results for each section are highlighted in \underline{underline}, respectively.}
    \label{tab:plm_tokenization}
\end{table*}

\subsection{Effect of Tokenization Granularity on \textbf{\texttt{SCRIPT}}}
\label{sec:plm_tokenization}
Beyond conducting ablations on individual components of the \textbf{\texttt{SCRIPT}} architecture (\S\ref{sec:ablation_study}), we also observed that the integration and effectiveness of compositional knowledge vary depending on the PLM's tokenization scheme. To investigate this, we compared four tokenization strategies: word, morpheme, subword, and character. As the final compositional token units changed accordingly, we also adjusted \textbf{\texttt{SCRIPT}}’s compressed output token unit to match.
Thus, instead of the original ``Character-to-Subword'' setting described in Section~\ref{sec:char_to_subword}, we experimented with ``Character-to-Word,'' ``Character-to-Morpheme,'' and ``Character-to-Character.''\footnote{To minimize OOV occurrences, we constructed vocabularies based on prior work~\cite{park-etal-2020-empirical, kim-etal-2024-kombo}, setting vocabulary sizes to 64k for Word, 32k for Morpheme and Subword, and 2k for Character.}

As shown in Table~\ref{tab:plm_tokenization}, \textbf{\texttt{SCRIPT}} proved effective when applied with larger units, such as Word and Morpheme, compared to Subword. This suggests that when the base PLM has already captured sufficient semantic meaning~\cite{aguilar-etal-2021-char2subword-extending, kaushal-mahowald-2022-tokens}, integrating syntactic compositional knowledge leads to a synergistic improvement. In contrast, with the smaller Character unit, where the base PLM primarily learns syntactic rather than semantic knowledge~\cite{aguilar-etal-2021-char2subword-extending, mielke2021between}, applying \textbf{\texttt{SCRIPT}} introduced noise and hindered performance.

\section{In-Depth Analysis}
Beyond the quantitative results, this section offers a detailed linguistic analysis of how \textbf{\texttt{SCRIPT}} operates in Korean. We examine how \textbf{\texttt{SCRIPT}} enriches subword representations and enables the base model to more effectively capture key linguistic phenomena.

\subsection{Impact on Morphological Variations}
\label{analysis:linguistic_phenomena}

\begin{figure}[h!]
  \centering
  \includegraphics[width=.9\linewidth]{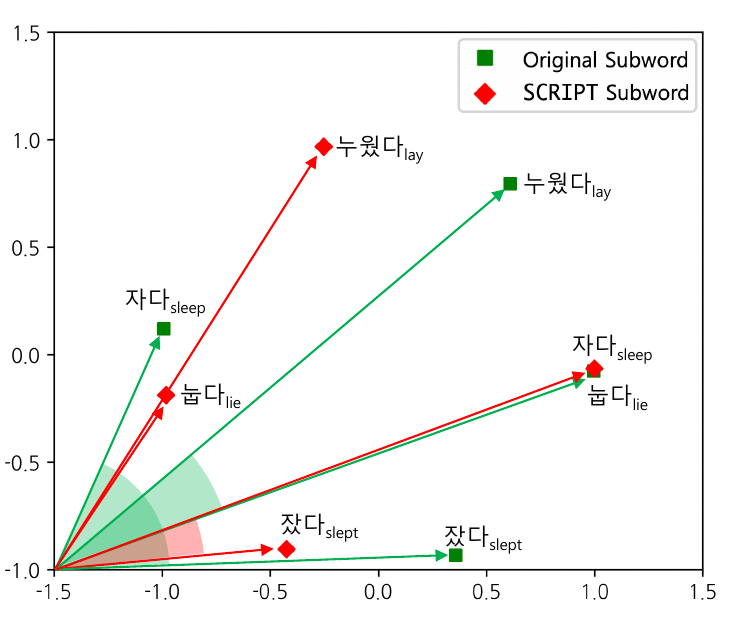}
  \caption{PCA visualization of subword embeddings for word pairs exhibiting subcharacter-level alternations. Each pair (e.g., 자다$_\text{sleep}$–잤다$_\text{slept}$, 눕다$_\text{lie}$–누웠다$_\text{lay}$) shares the same root meaning but differs in tense.}
  \label{fig:tense_inflections}
  \vspace{-1px}
\end{figure}

To assess how well \textbf{\texttt{SCRIPT}} captures subcharacter-level morphological alternations, we compare two subword representations: one from the PLM’s original subword embeddings and the other from \textbf{\texttt{SCRIPT}}’s subcharacter-based representations. Using these, we represent morphologically related word pairs, such as tense-inflected forms. Figure~\ref{fig:tense_inflections} provides a qualitative geometric illustration using mean-centered embeddings projected via PCA. In the projected space, \textbf{\texttt{SCRIPT}} places morphologically related forms in closer angular proximity, indicating a more structured encoding of tense relationships, while the original PLM embeddings appear more dispersed. To verify that this pattern is not an artifact of 2D projection, we additionally compute cosine similarity in the original embedding space over 50 verb–past tense pairs, observing a consistent increase from 0.71 to 0.80 (+11\%). These results suggest that subcharacter compositionality improves the model’s ability to capture fine-grained grammatical variations in Korean.

\begin{figure*}[t!]
  \centering
  \includegraphics[width=.99\linewidth]{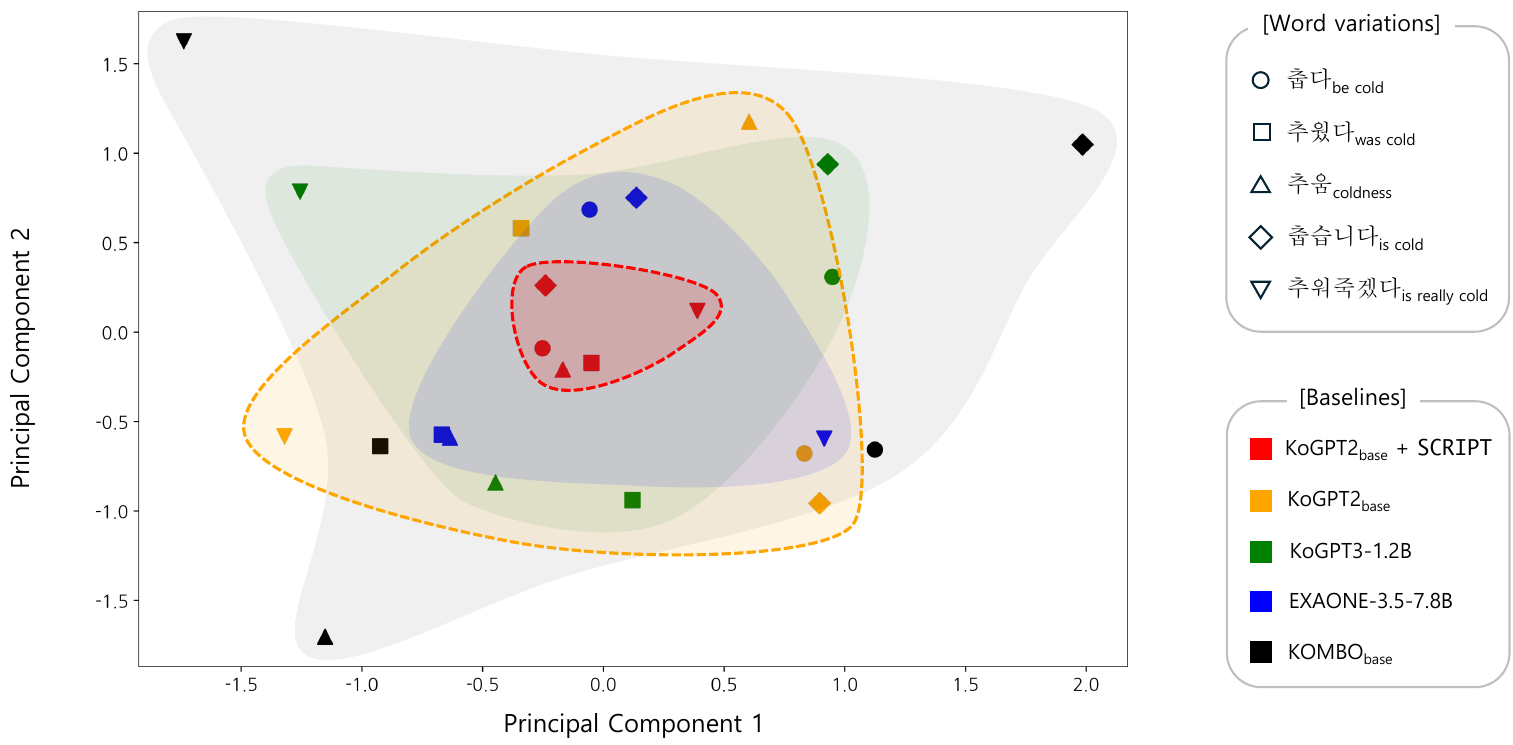}
  \caption{PCA visualization of word embeddings averaged over tokens for five semantically related Korean predicate inflections derived from the predicate `춥다$_\text{be cold}$'. The dashed boundaries indicate the dispersion ranges of the smallest baseline, KoGPT2$_\text{base}$, and its \textbf{\texttt{SCRIPT}}-augmented counterpart.}
  \label{fig:word_clustering}
\end{figure*}

\subsection{Impact on Word Embedding Cohesion}
As shown in Figure~\ref{fig:word_clustering}, we examine how Korean LMs organize semantically related predicate inflections in embedding space using five forms of the predicate `춥다$_\text{be cold}$'. Larger subword-based LMs show increasingly cohesive clustering, while the Jamo-based model, KOMBO~\cite{kim-etal-2024-kombo}, exhibits a more scattered distribution, likely due to its extreme focus on syntactic granularity. For clarity, we visualize KoGPT2 as a representative backbone, where \textbf{\texttt{SCRIPT}} produces the most compact grouping even at a small scale. Consistent patterns are observed across larger backbones. Together with ablation results showing degraded cohesion when compositional encoding is removed or altered, this suggests that the observed structure primarily arises from subcharacter compositional modeling rather than normalization effects.

\section{Related Work}
\label{section_2}

\subsection{Korean Pre-trained Language Models}
Most off-the-shelf Korean PLMs employ subword-based tokenization~\cite{yoo2024hyperclovaxtechnicalreport, exaone35technicalreport}, which has proven effective for handling Korean’s rich morphology.
However, such subword-based tokenizations do not explicitly model subcharacter-level structure, where many morphophonological processes in Korean occur.
To address this limitation, several studies have explored Jamo-level modeling of Korean~\cite{moon-okazaki-2020-jamo, cognetta-etal-2023-parameter, kim-etal-2024-kombo}.
In particular, \citet{kim-etal-2024-kombo} explicitly encodes the compositional structure of \textit{Hangul} to enrich character representations.
However, these approaches typically rely on non-standard or encoder-only architectures and require full pre-training from scratch, which limits their applicability to general-purpose PLMs.

\subsection{Multi-Granular Representations}
Some prior work has attempted to incorporate multiple granularities in Korean language modeling, such as combining Jamo and word embeddings~\cite{kwon-etal-2021-handling} or switching between Jamo and subwords depending on context~\cite{lee-etal-2025-jamo}. However, these methods typically alternate between token levels rather than structurally integrating them. Moreover, they often lack architectural generality or remain limited to encoder-based tasks. Despite growing evidence that combining fine-grained morphological cues with higher-level representations improves performance~\cite{lai-etal-2021-lattice, zhao2023mcl, wang-etal-2024-learning-mutually}, Korean PLMs still underexplore this integration in a principled and efficient manner.

\section{Conclusion}
This work presents \textbf{\texttt{SCRIPT}}, a modular framework for injecting subcharacter compositional knowledge into Korean PLM. Through a structure-aware compression mechanism grounded in the compositional principles of \textit{Hangul}, \textbf{\texttt{SCRIPT}} captures morphophonological variations at the subcharacter-level and enriches coarse PLM's token representations. Our experiments demonstrate that \textbf{\texttt{SCRIPT}} generally improves model performance across a wide range of Korean NLU and NLG tasks, enriching both conventional subword- and subcharacter-based approaches. 
Beyond quantitative gains, our linguistic analyses show that \textbf{\texttt{SCRIPT}} enhances the semantic and grammatical organization of the embedding space, enabling more cohesive clustering of inflected predicates and more faithful modeling of Korean linguistic phenomena. These findings underscore the limitations of subword tokenization in morphologically rich language and advocate for subcharacter-aware modeling as a necessary extension for Korean NLP.

\section*{Limitations}

This work focuses on improving Korean language understanding by explicitly modeling structural characteristics specific to Korean. Accordingly, \textbf{\texttt{SCRIPT}} is primarily designed and evaluated within the Korean linguistic context, and its effectiveness for other languages is not systematically validated in this study. Although the modular design may be adaptable to languages with rich internal character structure or complex morphology, such extensions are beyond the scope of this paper. As a minimal proof of concept, we show that \textbf{\texttt{SCRIPT}} can be integrated into a multilingual pre-trained model and still improves Korean task performance (Appendix~\ref{app:multilingual}). However, this experiment does not establish general cross-lingual applicability.

Second, while \textbf{\texttt{SCRIPT}} demonstrates consistent improvements across models ranging from approximately 100M to 2.4B parameters, we do not evaluate models at larger scales (e.g., 7B+). As larger models develop stronger internal representations, the relative benefit of explicitly modeling subcharacter structure may vary. Evaluating \textbf{\texttt{SCRIPT}} on larger-scale models remains an important direction for future work.

Finally, \textbf{\texttt{SCRIPT}} introduces additional parameters and sequence-length-dependent computations in the embedding layer, particularly when using cross-attention. Although this leads to improved convergence and performance, it also increases inference cost. Future work may explore more lightweight variants that better balance efficiency and effectiveness.

\section*{Acknowledgments}
This work was supported by the National Research Foundation of Korea (NRF) grant funded by the Korea government (MSIT) (No.RS-2025-00517221 and No.RS-2024-00415812) and Institute of Information \& communications Technology Planning \& Evaluation (IITP) grant funded by the Korea government (MSIT) (No.RS-2024-00439328, Karma: Towards Knowledge Augmentation for Complex Reasoning (SW Starlab), No.RS-2024-00457882, AI Research Hub Project, and No.RS-2019-II190079, Artificial Intelligence Graduate School Program (Korea University)).

\bibliography{custom}

\clearpage
\appendix

\begin{figure*}[t!]
    \centering
    \includegraphics[width=1.\linewidth]{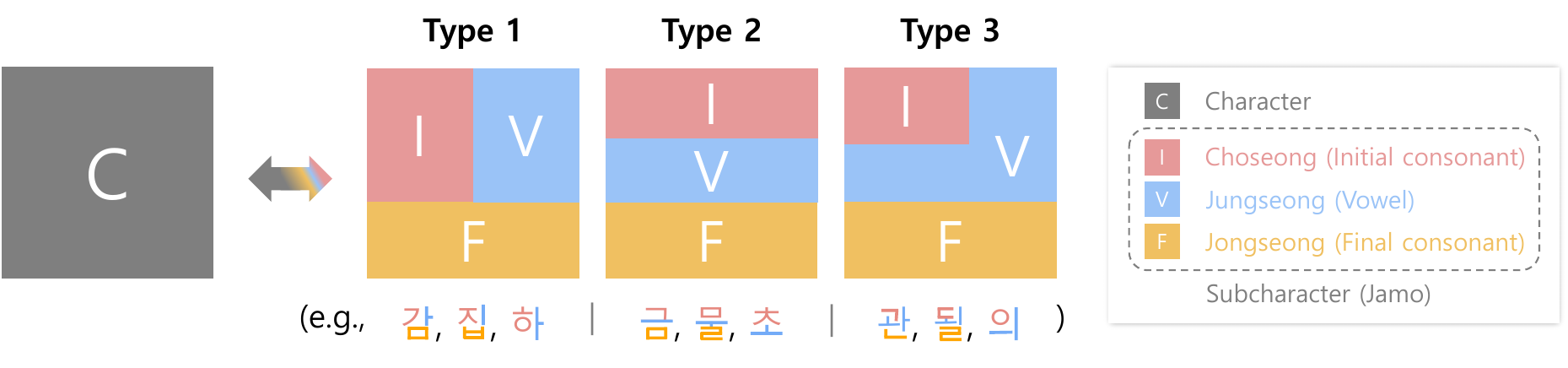}
    \caption{Three structural types of Korean syllable blocks, classified by the spatial arrangement of Choseong, Jungseong, and Jongseong.}
    \label{fig:character_structure}
\end{figure*}

\section{Composition of \textit{Hangul} Characters}
\label{app:hangul_character_explanation}

\textit{Hangul} characters are constructed by combining initial consonants (Choseong), medial vowels (Jungseong), and, optionally, final consonants (Jongseong). The following outlines the possible components for each position:

\subsection*{Initial Consonants (Choseong)}
The initial position can be occupied by one of the following 19 consonants:

\begin{center}
\small
\renewcommand{\arraystretch}{1.2}
\setlength{\tabcolsep}{5.8pt}
\begin{tabular}{ccccccccccc}
    ㄱ & ㄲ & ㄴ & ㄷ & ㄸ & ㄹ & ㅁ & ㅂ & ㅃ & ㅅ & ㅆ \\
    ㅇ & ㅈ & ㅉ & ㅊ & ㅋ & ㅌ & ㅍ & ㅎ & & &
    \end{tabular}
    \newline
\end{center}

\subsection*{Medial Vowels (Jungseong)}
The medial position consists of 21 vowels, which can be categorized based on their placement relative to the initial consonant:

\begin{itemize}
  \item \textbf{Vertical vowels} (placed to the right of the initial consonant): ㅏ, ㅑ, ㅓ, ㅕ, ㅣ, ㅐ, ㅒ, ㅔ, ㅖ
  \item \textbf{Horizontal vowels} (placed below the initial consonant): ㅗ, ㅛ, ㅜ, ㅠ, ㅡ
  \item \textbf{Complex vowels} (combining both vertical and horizontal elements): ㅘ, ㅙ, ㅚ, ㅝ, ㅞ, ㅟ, ㅢ
\end{itemize}
These three categories correspond to the examples illustrated in Figure~\ref{fig:character_structure}. This classification results in three distinct types of character shapes, each reflecting the spatial arrangement dictated by the vowel's orientation.
\label{app:vowel_explanation}

\subsection*{Final Consonants (Jongseong)}
The final position may be unoccupied or contain one of the following 27 consonant combinations. {\tiny \color{gray} \pmboxdrawuni{2583}} means the empty final consonant and it is also considered a valid configuration:

\begin{center}
    \small
    \renewcommand{\arraystretch}{1.2}
    \setlength{\tabcolsep}{5.8pt}
    \begin{tabular}{ccccccccccc}
        ㄱ & ㄲ & ㄳ & ㄴ & ㄵ & ㄶ & ㄷ & ㄹ & ㄺ & ㄻ \\
        ㄼ & ㄽ & ㄾ & ㄿ & ㅀ & ㅁ & ㅂ & ㅄ & ㅅ & ㅆ \\
        ㅇ & ㅈ & ㅊ & ㅋ & ㅌ & ㅍ & ㅎ & {\tiny \color{gray} \pmboxdrawuni{2583}} & & &
    \end{tabular}
    \newline
\end{center}

\subsection*{Syllable Structure}
A typical \textit{Hangul} syllable block is formed in one of the following structures:

\begin{itemize} [leftmargin=10px]
  \item \textbf{CV}: Consonant + Vowel (e.g., 가)
  \item \textbf{CVC}: Consonant + Vowel + Consonant (e.g., 갈)
\end{itemize}
The positioning of vowels within the syllable block depends on their type:

\begin{itemize}
  \item \textbf{Vertical vowels} are placed to the right of the initial consonant.
  \item \textbf{Horizontal vowels} are placed below the initial consonant.
  \item \textbf{Complex vowels} may occupy both right and bottom positions relative to the initial consonant.
\end{itemize}
This systematic arrangement allows for the construction of 11,172 possible syllable combinations in \textit{Hangul}.

\section{Details for Inflection Frequency Evaluation}
\label{app:inflection_freq}

\paragraph{Action Definition.}
The alignment oracle~\cite{matteson-etal-2018-rich} aligns surface forms with lemma sequences by assigning one or more actions to each input character:
\begin{itemize}
    \item \textbf{KEEP}: retain the character unchanged.
    \item \textbf{MOD}: modify the character into a different form (e.g., adding a final consonant, changing a vowel, or altering the initial consonant).
    \item \textbf{NOOP}: drop the character, i.e., it does not appear in the lemma.
\end{itemize}
Each action is augmented with BIO prefixes: ``B-'' marks the beginning of a morpheme, while ``I-'' denotes continuation within a morpheme. For instance, \texttt{B-KEEP} indicates the start of a morpheme where the character is preserved, while \texttt{B-MOD-ㄴ} signals a morpheme-internal modification introducing the consonant ``ㄴ.''

\begin{table*}[h!]
    \renewcommand{\arraystretch}{1.2}
    \centering
    \begin{tabular}{l|l|l}
        \hline
        \textbf{Input Character} & \textbf{Oracle Actions} & \textbf{Output Lemma Units} \\
        \hline
        런 (reon) & B-MOD-럽, I-MOD-ㄴ & 럽 (reob), ㄴ (n) \\
        했 (haess) & B-MOD-하, I-MOD-았 & 하 (ha), 았 (ass) \\
        다 (da) & B-KEEP & 다 (da) \\
        \hline
    \end{tabular}
    \caption{Examples of oracle actions aligned with lemma units. Each input character may correspond to multiple oracle actions (e.g., B-MOD, I-MOD) depending on the type and structure of the morphological transformation. The output shows the resulting lemma units produced by these actions.}
    \label{tab:multi_action}
\end{table*}

\paragraph{Corpus Preprocessing.}  
We first parse an oracle-aligned action file in which each line contains a single input character and its action sequence (possibly multiple actions for one character), separated by a fixed delimiter. Non-\textit{Hangul} characters are filtered out using a \textit{Hangul} checker (we use \texttt{hgtk.checker.is\_hangul~\footnote{https://pypi.org/project/hgtk/}}). For each Korean character, we collect the raw action string and increment counters for \texttt{KEEP}, \texttt{MOD}, or \texttt{NOOP} depending on whether the action string contains these tokens. BIO prefixes (\texttt{B-}/\texttt{I-}) are preserved to indicate morpheme boundaries.~\cite{matteson-etal-2018-rich}


\paragraph{Why Focus on MOD.}
Our analysis concentrated on ``MOD'' actions because they directly represent the sites of morphophonological change. ``KEEP'' characters reflect unchanged segments and ``NOOP'' characters denote deletions, both of which contribute little information about how Korean morphology operates. In contrast, ``MOD'' characters capture precisely the subcharacter or character-level transformations (e.g., tense, honorifics, adnominal endings) that are central to Korean grammar. By quantifying only ``MOD,'' we obtain a clearer picture of where and how morphophonological alternations occur.

\paragraph{Granularity Classification (Subcharacter vs Character).}
We classify each \texttt{MOD} instance into two levels:
\begin{itemize}[leftmargin=15px]
    \item \textbf{Subcharacter-level}: exactly one of \{Choseong (I), Jungseong (V), Jongseong (F)\} differs between the input and the aligned output, or a cross-syllable transfer/merge occurs, consistent with Korean fusion rules across character boundaries.~\cite{matteson-etal-2018-rich}
    \item \textbf{Character-level}: entire syllable box is replaced wholesale (non-comparable at the subcharacter-level).
\end{itemize}
In our implementation, we filter Korean characters (\texttt{hgtk.checker.is\_hangul}) and compute counts of \texttt{KEEP}/\texttt{MOD}/\texttt{NOOP}. For \texttt{MOD} cases, we determine the granularity by comparing the input character’s Unicode-decomposed (I, V, F) tuple with that of its aligned outputs. When a character yields multiple actions, we first reconstruct the immediate output units produced by that character’s actions and then compare at the subcharacter-level. If only one subcharacter differs (or a Jongseong-Choseong transfer is observed), we mark it as subcharacter-level; otherwise, character-level.

\paragraph{NOOP Handling.}
\texttt{NOOP} marks deletions (input characters with no aligned lemma). As deletions indicate absence rather than transformation, we exclude \texttt{NOOP} from granularity statistics; counts are still reported for completeness. (Alternatively, \texttt{NOOP} can be treated as character-level; we opted for exclusion.)


\begin{table*}[t!]
    \centering
    \begin{tabular}{@{}l@{}}
        \toprule
        \textbf{Algorithm 1: Subword Representation with \textbf{\texttt{SCRIPT}}}\\
        \midrule
        \textit{Input:}\; $\mathbf{s}$ (raw input sentence), embedding dimension $D$.\\
        \textit{Output:}\; $\mathbf{e}_\text{F}$ (fused subword-level embeddings).\\\\[2pt]
        
        \textbf{1}\; $\mathbf{t}_\text{subchar} \leftarrow \textsc{Tokenize}_\text{subchar}(\mathbf{s})$ \hspace*{2.2cm} // subcharacter sequence\\
        \textbf{2}\; $\mathbf{e}\leftarrow\textsc{Embedding}_\text{subchar}(\mathbf{t}_\text{subchar})$ \hspace*{1.8cm} // subcharacter representation\\
        \textbf{3}\; \textbf{if} first token in $\mathbf{e}$ is \texttt{[CLS]} \textbf{then} \hspace*{2.1cm} // (for encoder-only model) \\
        \hspace*{2.6em}$\mathbf{e}_{\text{CLS}} \leftarrow \mathbf{e}[1]$\\
        \hspace*{4.0em}$\mathbf{e} \leftarrow \mathbf{e}[2:N]$\\
        \textbf{4}\; $\mathbf{h} \leftarrow \textsc{GRU}(\mathbf{e})$ \hspace*{4.8cm} // sequential composition \\
        \textbf{5}\; \textbf{for} each for each integer $k$ in [1, N/3]: \\
        \hspace*{2.8em} $\mathbf{h}_\text{I} \leftarrow \mathbf{h}[3k-2]$ \hspace*{3.7cm}\, // \text{Choseong} \\
        \hspace*{2.6em} $\mathbf{h}_\text{V} \leftarrow \mathbf{h}[3k-1]$ \hspace*{3.7cm}\, // \text{Jungseong} \\
        \hspace*{2.8em} $\mathbf{h}_\text{F} \leftarrow \mathbf{h}[3k]$ \hspace*{4.3cm} // \text{Jongseong} \\
        \textbf{6}\; $\mathbf{h}_\text{I+V}\leftarrow\textsc{GRU}(\mathbf{h}_I\!+\!\mathbf{h}_V)$ \hfill \hspace*{3.5cm} // fusion of Choseong and Jungseong\\
        \textbf{7}\; $\mathbf{h}_\text{R}\leftarrow\textsc{Stack}(\mathbf{h}_I\!+\!\mathbf{h}_V,\mathbf{h}_F)$ \hspace*{2.9cm} // reshape subcharacter sequence\\
        \textbf{8}\; $\mathbf{h}_\text{C}\leftarrow\textsc{AvgPool}(\textsc{Conv}_{2\times1}(\mathbf{h}_\text{R}))$ \hspace*{1.8cm}  // character representation \\
        \textbf{9}\; $\mathbf{h}_\text{S}\leftarrow\textsc{Pooling}(\textsc{GRU}(\mathbf{h}_\text{C}))$ \hspace*{2.6cm} // subword representation of \textbf{\texttt{SCRIPT}} \\\\
        
        \textbf{10}\; $\mathbf{t}_\text{subword} \leftarrow \textsc{Tokenize}_\text{subword}(\mathbf{s})$ \hspace{2.0cm} // subword sequence\\
        \textbf{11}\; $\mathbf{e}_\text{S} \leftarrow \textsc{Embedding}_\text{subword}(\mathbf{t}_\text{subword})$ \hspace*{1.4cm} // original subword representation \\\\
        
        \textbf{12}\; $\mathbf{e}_\text{F} \leftarrow$ \textsc{CrossAttn}(Q=$\mathbf{e}_\text{S}$,\,K=$\mathbf{h}_\text{S}$,\,V=$\mathbf{h}_\text{S}$) \hspace*{0.7cm} // fusion of subword representations \\
        \textbf{13}\; \textbf{if} $\mathbf{e}_{\text{CLS}}$ exists \textbf{then} $\mathbf{e}_\text{F} \leftarrow [\mathbf{e}_{\text{CLS}}; \mathbf{e}_\text{F}]$ \\\\
        
        \textbf{return} $\mathbf{e}_\text{F}$\\
        \bottomrule
    \end{tabular}
    \caption{The input text $\mathbf{s}$ is tokenized into subcharacter tokens (line 1) and subword tokens (line 10), respectively (\S~\ref{sec:script_architecture}). While structure-aware subword embeddings are built bottom-up from subcharacter to character to subword (lines 2-9), the original subword embeddings are looked up (lines 10-11). These two subword representation streams are fused by cross-attention to yield the final subword representation $\mathbf{e}_\text{F}$ (lines 12-13) (\S~\ref{sec:fusion_subword}).}
    \label{tab:script_algo}
\end{table*}

\begin{figure*}[t!]
    \centering
    \includegraphics[width=1.\linewidth]{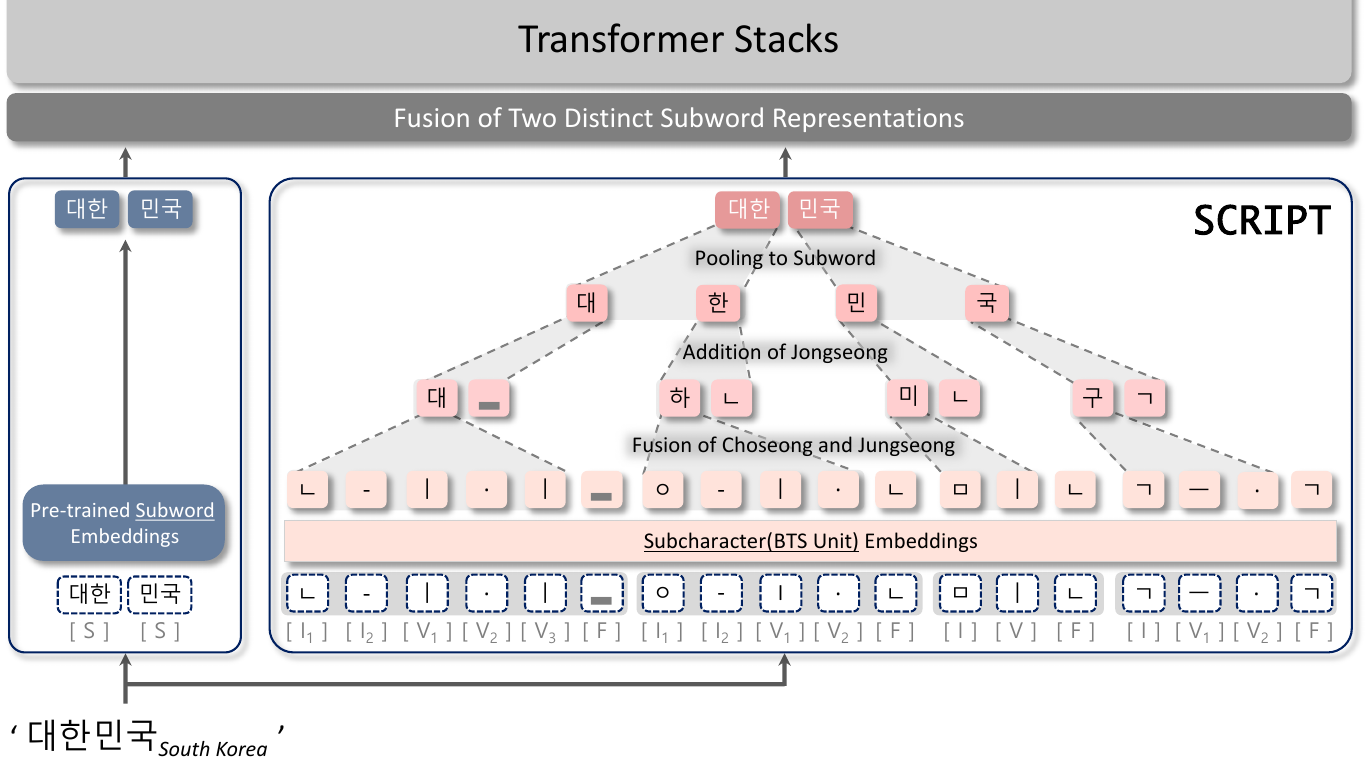}
    \caption{
    Implementation of \textbf{\texttt{SCRIPT}} for the BTS unit. This illustrates the hierarchical integration of subword representations derived from BTS unit in \textbf{\texttt{SCRIPT}}, using the example word `대한민국\textit{$_\text{South Korea}$}'. The word `대한민국\textit{$_\text{South Korea}$}' consists of two subwords ([S]: 대한, 민국), four characters (대, 한, 민, 국), and eighteen subcharacters: initial consonants ([\,I\,]: ㄴ, -, ㅇ, -, ㅁ, ㄱ), vowels ([V]: ㅣ, \,\,$\cdot$\,, ㅣ, ㅣ, \,\,$\cdot$\,, ㅣ, ㅡ, \,\,$\cdot$\,), and final consonants ([F]: {\tiny \color{gray} \pmboxdrawuni{2583}}\;, ㄴ, ㄴ, ㄱ).
    }
    \label{fig:bts_script_architecture}
\end{figure*}

\section{Implementation Details and Model Considerations}
\label{app:model_consideration}
As shown in Figure~\ref{fig:script_details}, our proposed method, \textbf{\texttt{SCRIPT}}, can be simply plugged into the embedding layer, making it easy to apply to existing PLMs. This provides a model-agnostic advantage, allowing seamless integration with various architectures. 
As an extra implementation detail, unlike the decoder model, such as GPT, the encoder model, like BERT, has a particular design of the input sequence. BERT adds a special token \texttt{[CLS]} at the beginning of the input sequence, and this token is used as the representation of the sequence at the last-layer hidden state. To preserve the special structure and meaning of the special token during the compression stage, we first separate the \texttt{[CLS]} token from the rest of the tokenized sequence (line 3 in Table~\ref{tab:script_algo}). Then, we only use the remaining subcharacter sequence as input to \textbf{\texttt{SCRIPT}}. After processing through \textbf{\texttt{SCRIPT}}, we add the \texttt{[CLS]} hidden state back to the output of \textbf{\texttt{SCRIPT}}, the compressed subword representation (line 13 in Table~\ref{tab:script_algo}).
Formally, this can be represented as $\mathbf{h}_\text{S} \in\mathbb{R}^{(N'+1) \times \text{D}}$:
\begin{equation} \label{eq:12}
    \mathbf{h}_\text{S} = \textsc{Emb}_\text{subchar}(\mathbf{t}_{0}) \oplus \textbf{\texttt{SCRIPT}}(\mathbf{t}_{2:N+1})
\end{equation}
This trivial technique provides the versatility for our proposed methodology, \textbf{\texttt{SCRIPT}}, to be easily applied across all existing PLMs. In the following experimental section, we demonstrate the utility of our methodology by applying \textbf{\texttt{SCRIPT}} to both pre-trained encoder-only and decoder-only models.

Another important consideration is the integration of the two distinct subword representations, which come from different levels of granularity. This process is highly sensitive to normalization, as it can easily disrupt compatibility between the pre-trained subword embeddings and those derived from subcharacter representations. Notably, our empirical analysis shows that the pre-trained embeddings are approximately 30 times larger in size than those generated by \textbf{\texttt{SCRIPT}}. Normalizing these vectors to a common scale can distort the distribution of the pre-trained embeddings, with the randomly initialized subcharacter representations introducing significant noise and resulting in the loss of crucial learned information.
Our analysis shows that structure-aware subword knowledge is gradually fine-tuned and transferred into the pre-trained embeddings during integration. Therefore, normalizing the two representations to the same scale before fusion risks compromising the effectiveness of the pre-trained subword embeddings; it is empirically more effective to use them as they are.

\section{Implementation Details for BTS Units}
\label{app:bts_units}
Unlike Section \ref{sec:script_architecture}, where we introduced the architecture of \textbf{\texttt{SCRIPT}} using Jamo as the base unit, this section extends the discussion to describe \textbf{\texttt{SCRIPT}} in terms of other subcharacters, such as BTS units. 
As noted in Section \ref{sec:methodology}, BTS units are alternative subcharacter representations of \textit{Hangul}, decomposing each character into even finer subcomponents than Jamo.
According to \citet{kim-etal-2022-break}, consonants can be split into up to four subcomponents (e.g., the consonant `ㅉ’ decomposes into \{ㅅ, -, ㅅ, -\;\}, while vowels can be split into up to five subcomponents (e.g., the vowel `ㅙ’ decomposes into \{\;$\cdot$\;, ㅡ,ㅣ, \;$\cdot$\;,ㅣ\}. Depending on the selective decomposition of consonants or vowels, there are three distinct types of units: consonant-only decomposition (denoted as Stroke), vowel-only decomposition (denoted as Cji, short for \textit{Cheonjiin}), and both consonant and vowel decomposition (denoted as BTS). As a result, the maximum number of tokens per character varies depending on the decomposition type: the Stroke decomposition results in up to 9 subcharacter tokens, the Cji decomposition yields up to 7 tokens, and the BTS decomposition produces up to 13 tokens.

To incorporate the information of BTS units into the original subword representation of PLMs, we follow the progressive steps outlined in Section~\ref{sec:script_architecture} in a similar manner. For simplicity, we employ BTS as the initial subcharacter unit of \textbf{\texttt{SCRIPT}} in the following explanation. 
We first tokenize the input text $\mathbf{s}$ into subcomponents, such as BTS, then project the resulting subcharacter sequence $\mathbf{t}_\text{subchar}$ into a subcharacter embedding space. A GRU layer is applied sequentially for contextualization:

\begin{equation} \label{eq:13}
  \mathbf{t}_\text{subchar} = \textsc{Tokenize}_\text{subchar}(\mathbf{s}) \quad \in\mathbb{R}^N
  \vspace{-0.7cm}
\end{equation}

\begin{equation} \label{eq:14}
  \mathbf{e} = \textsc{Emb}_\text{subchar}(\mathbf{t}) \quad \in \mathbb{R}^{N \times D}
  \vspace{-0.7cm}
\end{equation}

\begin{equation} \label{eq:15}
    \mathbf{h} = \textsc{GRU}(\mathbf{e}) \quad \in\mathbb{R}^{N \times D}
\end{equation}
Next, we merge the subcomponent tokens to construct representations for Choseong, Jungseong, and Jongseong. For each integer $k \in [1, N/13]$, the representations of Choseong ($h_{\text{I},k}$), Jungseong ($h_{\text{V},k}$), and Jongseong ($h_{\text{F},k}$) are defined as follows:
\begin{align}
    \mathbf{h}_{\text{I}, k} &= \sum_{j=1}^{4}\mathbf{h}_{13(k-1) + j} \quad \in \mathbb{R}^{\frac{N}{13}\times D} \label{eq:16} \\
    \mathbf{h}_{\text{V}, k} &= \sum_{j=5}^{9}\mathbf{h}_{13(k-1) + j} \quad \in \mathbb{R}^{\frac{N}{13}\times D} \label{eq:17} \\
    \mathbf{h}_{\text{F}, k} &= \sum_{j=10}^{13}\mathbf{h}_{13(k-1) + j} \quad \in \mathbb{R}^{\frac{N}{13}\times D} \label{eq:18}
\end{align}
Next, we combine the representation of Choseong $\mathbf{h}_\text{I}$ and Jungseong $\mathbf{h}_\text{V}$:
\begin{equation}
\label{eq:19} 
    \mathbf{h}_\text{I+V}=\mathbf{h}_\text{I}+\mathbf{h}_\text{V} \quad \in \mathbb{R}^{\frac{N}{13}\times D} 
\end{equation}
After that, we vertically concatenate the combined representation $\mathbf{h}_\text{I+V}$ with Jongseong representation $\mathbf{h}_\text{F}$:
\begin{equation}
\label{eq:20} 
    \mathbf{h}_\text{R}= 
        \begin{bmatrix} 
        \mathbf{h}_\text{I+V} \\
        \mathbf{h}_\text{F} 
        \end{bmatrix} 
    \quad \in\mathbb{R}^{2\times{\frac{N}{13}}\times D} 
\end{equation}
To generate the dense character representation, we merge these vertically aligned representations by applying a convolution and a pooling layer:
\begin{equation}
\label{eq:21} 
    \mathbf{h}_\text{C}= \textsc{AvgPool}(\textsc{Conv}(\mathbf{h}_\text{R})) \quad \in\mathbb{R}^{\frac{N}{13}\times D} 
\end{equation}

Finally, a GRU layer is applied to the character representations $\textbf{h}_\text{C}$, followed by a pooling layer to compress and align the granularity of the character representations with that of the original subword representations:
\begin{equation} 
\label{eq:22} 
    \mathbf{h}_\text{S} = \textsc{Pooling}(\textsc{GRU}(\mathbf{h}_\text{C})) \quad \in\mathbb{R}^{N' \times D} 
\end{equation}
where $N'$ represents the number of tokens of the original subword sequence. This compressed subword representation $\mathbf{h}_\text{S}$ is fused with the original subword representation $\mathbf{e_{S}}$ through cross-attention, as described in Section~\ref{sec:fusion_subword}, to produce the final subword representation.

\section{Experimental Settings}
\label{app:experimental_settings}

\subsection{Baselines}
\label{app:baselines}
To evaluate the effectiveness of our proposed method, \textbf{\texttt{SCRIPT}}, we apply it to various PLMs listed below. 
Models equipped with our method are denoted as `\texttt{model}+\textbf{\texttt{SCRIPT}}'. When needed, the subcharacter type (e.g., Jamo) used in \textbf{\texttt{SCRIPT}} is indicated as a subscript, as in \textbf{\texttt{SCRIPT}}$_\text{Jamo}$. If no subscript is provided, Jamo is used by default.
\begin{itemize}
    \item BERT$_\text{base}$ \cite{devlin2019bert}: A bidirectional language model based on the Transformer architecture. It consists of multiple Transformer encoder layers. It is pre-trained in a self-supervised manner, enabling it to learn without labeled data. We utilize the BERT$_\text{base}$ model, which includes 12 Transformer encoder layers. It has a total of 110 million parameters. We employ a morpheme-aware tokenizer~\cite{park-etal-2020-empirical} with a vocabulary size of 32k. We pre-train the BERT$_\text{base}$ model for 1 million steps on Masked Language Modeling (MLM) and Next Sentence Prediction (NSP) tasks, using a corpus of 6.2GB consisting of the Korean Wikipedia and Namuwiki.~\footnote{\url{https://namu.wiki/}}
    
    \item KOMBO$_\text{base}$ \cite{kim-etal-2024-kombo}: A Jamo-based Korean encoder-only PLM that leverages the invention principles of \textit{Hangul} to represent characters. While the architecture of KOMBO$_\text{base}$ is designed based on BERT$_\text{base}$, it differs in that it uses subcharacter-level tokens instead of subwords. It also includes a combination layer below and a restoration layer above its 12 Transformer blocks, introducing additional computational cost and overhead. However, it achieves better performance on NLU tasks than BERT$_\text{base}$. Moreover, since both BERT$_\text{base}$ and KOMBO$_\text{base}$ models are encoder-only models, they do not apply to generative tasks. We pre-train KOMBO$_\text{base}$ for 1 million steps on MLM and NSP tasks using a 6.2GB corpus from Korean Wikipedia and Namuwiki, following the same configuration as BERT$_\text{base}$.
    
    \item KoGPT2$_\text{base}$~\footnote{\url{https://github.com/SKT-AI/KoGPT2}}: A generative language model composed of multiple Transformer decoder blocks. Unlike the BERT model, which is trained on MLM and NSP tasks, GPT is trained on a next token prediction task, enabling it to generate contextually relevant text. KoGPT2 is a Korean variant of the GPT model, following the GPT2$_\text{base}$ configuration with 12 Transformer decoder blocks and 125 million parameters. It has been pre-trained on the Korean Wiki and Korpora datasets, totally over 40GB. KoGPT2 employs subword tokenization with a vocabulary size of 51.2k.
    \item KoGPT3-1.2B~\footnote{\url{https://huggingface.co/skt/ko-gpt-trinity-1.2B-v0.5}}: A large-scale Transformer decoder model containing 1.2 billion parameters and 24 Transformer decoder blocks. This model follows the GPT-3 architecture. The model is trained on Ko-DAT, a large-scale, curated Korean dataset created by SK Telecom with 35 billion tokens, using the next token prediction task over 72k training steps. Similar to KoGPT2$_\text{base}$, KoGPT3-1.2B uses subword tokenization with a 51.2k vocabulary size.
    \item mGPT-1.3B \cite{shliazhko-etal-2024-mgpt}: The multilingual extension of GPT-3 which is pre-trained across 61 languages. It consists of 24 Transformer decoder layers with a total of 1.3 billion parameters. Pre-training was conducted on the Wikipedia and C4 corpora, over 600GB in total, for 600k steps. The model employs a 100k size vocabulary and utilizes Byte-level Byte Pair Encoding (BBPE) as its default tokenization strategy, enhancing its multilingual capabilities.
    \item EXAONE-3.5-2.4B-Instruct (EXAONE-2.4B for short) \cite{exaone35technicalreport}: A bilingual (Korean and English) instruction-tuned language model developed by LG AI Research. It uses a decoder-only Transformer architecture and is part of the EXAONE 3.5 series. The model has 30 Transformer decoder layers and 2.41 billion parameters total. It was trained under a causal/next-token prediction objective using a bilingual corpus curated by LG AI Research, supports a maximum context length of 32,768 tokens, and employs a shared vocabulary of 102,400 tokens using a BBPE tokenizer.
\end{itemize}

\subsection{Implementation Details}
\label{app:implementation_details}
We utilized a series of pre-trained GPT models, including KoGPT2$_\text{base}$, KoGPT3-1.2B, mGPT-1.3B, and EXAONE-2.4B all sourced from the Huggingface library\footnote{\url{https://huggingface.co/models}}, and fine-tuned them using LoRA (Low-Rank Adaptation)~\cite{hu2021lora}. However, for BERT-based models, such as BERT$_\text{base}$ and KOMBO$_\text{base}$, training with LoRA showed instability, so we opted for full fine-tuning exclusively for these models. As noted by \citet{hu2021lora}, there was no significant difference in performance between LoRA and full fine-tuning. Since we utilize the DeepSpeed library\footnote{\url{https://github.com/deepspeedai/DeepSpeed}} for models larger than 1B parameters, all models were trained on a single NVIDIA RTX 3090 GPU.
We set the default maximum sequence length to 256 for the original PLM's subword tokenizer, and to 2048 for the subcharacter tokenizer used in KOMBO$_\text{base}$ and \textbf{\texttt{SCRIPT}}$_\text{Jamo}$. For tasks requiring longer inputs, such as HellaSwag and XL-Sum, we used sequence lengths of 512 and 3072, respectively. We use the \texttt{AdamW} optimizer and cosine learning rate scheduler. For most other experimental settings, we used the default configurations of each pre-trained model. The detailed hyperparameter settings are summarized in Table~\ref{tab:hyperparameters}. All experiments are repeated over 3 random seeds (42--44), and we report the mean.

\begin{table*}[t!]
  \centering
  \small
  \setlength{\tabcolsep}{5pt}         
  \renewcommand{\arraystretch}{1}    
  \begin{tabular}{l|ccccccc}
    \toprule
    \multirow{2}{*}{Task} & \multirow{2}{*}{Epoch} & \multirow{2}{*}{\shortstack{Batch\\Size}} & \multirow{2}{*}{\shortstack{Learning\\Rate}} & \multirow{2}{*}{\shortstack{Dropout\\Ratio}} & \multirow{2}{*}{\shortstack{Warmup\\Ratio}} & \multirow{2}{*}{LoRA r} & \multirow{2}{*}{LoRA $\alpha$} \\ \\
    \midrule
    KorNLI      & 5    & 64  & \multirow{4}{*}{\shortstack[l]{
    BERT: \{5e-05, 1e-04\} \\
    \shortstack{GPT\;\:: \{1e-05, 5e-05, 1e-04,\\ \quad\quad\quad 1e-03, 3e-03, 1e-02\}}
    }} & \multirow{4}{*}{0.03} & \multirow{4}{*}{0.1} & \multirow{4}{*}{32} & \multirow{4}{*}{128}  \\
    KorSTS      & 15   & 64  &  &  &  &  &  \\
    NSMC        & 5    & 64  &  &  &  &  &  \\
    PAWS-X      & 10   & 64  &  &  &  &  &  \\
    \midrule
    BoolQ      & 10   & 8  & \multirow{5}{*}{\shortstack[l]{
    BERT: \{1e-05, 5e-05\} \\
    \shortstack{GPT\;\:: \{1e-05, 5e-05, 1e-04,\\ \quad\quad\quad 1e-03, 3e-03, 1e-02\}}
    }} & \multirow{5}{*}{0.03} & \multirow{5}{*}{0.1} & \multirow{5}{*}{32} & \multirow{5}{*}{128} \\
    COPA       & 15   & 16  &  &  &  &  &  \\
    WiC        & 15   & 16  &  &  &  &  &  \\
    HellaSwag  & 10   & 8   &  &  &  &  &  \\
    SentiNeg   & 10   & 64  &  &  &  &  &  \\
    \midrule
    KoCommonGen & 15   & 64  & \multirow{5}{*}{
    \shortstack{GPT\;\:: \{1e-04, 1e-03, 1e-02, \\ \quad\quad\quad 2e-02, 3e-02, 4e-02\}}
    } & \multirow{5}{*}{0.03} & \multirow{5}{*}{0.1} & \multirow{5}{*}{32} & \multirow{5}{*}{128}  \\
    XL-Sum      & 10   & 64  &  &  &  &  &  \\
    Kor-Learner & 10   & 64  &  &  &  &  &  \\
    Kor-Native  & 10   & 64  &  &  &  &  &  \\
    \bottomrule
  \end{tabular}
  \caption{Hyperparameters used in all experiments in this paper for each task. For the learning rate, we select the value that yields the best performance for each baseline. Here, ``BERT'' refers to encoder-only models, including BERT and KOMBO, while ``GPT'' encompasses all decoder-only models, such as KoGPT2, KoGPT3, mGPT, and EXAONE.}
  \label{tab:hyperparameters}
\end{table*}

\subsection{Tasks}
\label{app:tasks}
\paragraph{Korean NLU Tasks.}
To investigate the performance of our proposed method on Korean NLU tasks, we evaluated baselines on nine distinct Korean NLU datasets. 
Four of these, KorNLI, KorSTS, NSMC, and PAWS-X, are widely used benchmarks for Korean NLU tasks, which we refer to as ``\textit{Korean Standard NLU Tasks}''~\cite{jang-etal-2022-kobest}. The remaining five datasets belong to the KoBEST benchmark (abbreviated as KB), which is designed to evaluate Korean language models on more complex linguistic understanding. We refer to these as ``\textit{Korean Advanced NLU Tasks}''~\cite{jang-etal-2022-kobest}.
\begin{itemize}
    \item KorNLI \cite{ham-etal-2020-kornli}: A dataset comprising 943k train, 25.5k validation, and 5k test samples for NLI, derived from the SNLI \cite{bowman-etal-2015-large}, MNLI \cite{williams-etal-2018-broad}, and XNLI \cite{conneau-etal-2018-xnli} datasets. The data is labeled across three classes: entailment, neutral, and contradiction.
    \item KorSTS \cite{ham-etal-2020-kornli}: A dataset developed to assess the semantic similarity between sentence pairs, adapted from the Korean STS-B dataset \cite{cer-etal-2017-semeval}. KorSTS consists of 5,749 train samples and 2,879 evaluation samples, each labeled with a similarity score from 0 to 5, indicating the degree of semantic similarity between the sentences.
    \item NSMC \cite{Park:2016}: A dataset sourced from NAVER is used for sentiment analysis of Korean movie reviews. It includes 150k train samples and 50k test samples, with each review labeled as either negative or positive.
    \item PAWS-X \cite{yang-etal-2019-paws}: A dataset for paraphrase identification, which includes six different language tasks. We only use the Korean subset to evaluate models. It contains 53k sentence pairs (49k for train, 2k for development, and 2k for test), each data labeled with one of two values: different meanings or paraphrases.
    \item KoBEST \cite{jang-etal-2022-kobest}: A benchmark suite designed to evaluate broad linguistic and cognitive capabilities of Korean language models through five diverse and challenging tasks.
    \begin{itemize} [leftmargin=15px]
        \item KB-BoolQ \cite{jang-etal-2022-kobest}: A dataset of 3.7k train, 700 validation, and 1.4k test instances. The task is a true/false question and answer format based on paragraphs, with sources from Korean Wikipedia.
        \item KB-COPA \cite{jang-etal-2022-kobest}: A dataset includes 3.1k train, 1k validation, and 1k test instances. Models predict cause or effect given a premise, designed similarly to the English COPA dataset \cite{roemmele2011choice}.
        \item KB-WiC \cite{jang-etal-2022-kobest}: A dataset contains 3.3k train, 1.3k validation, and 1.3k test samples, requiring models to determine if a target word holds the same meaning across two contexts.
        \item KB-HellaSwag \cite{jang-etal-2022-kobest}: A dataset composed of 2k train, 500 validation, and 500 test examples, where models select the most probable sentence to follow a given context. The data is sourced from YouTube and Wikipedia.
        \item KB-SentiNeg \cite{jang-etal-2022-kobest}: A dataset for sentiment analysis (3.6k train, 400 validation, 397 test samples) focusing on the polarity of negated sentences in product reviews.
    \end{itemize}
\end{itemize}

\begin{table*}[t!]
  \centering
  \small
  \setlength{\tabcolsep}{5pt}       
  \renewcommand{\arraystretch}{1.}    
  \begin{tabular}{lccccccc}
    \toprule
    \multirow{2.5}{*}{Model} & \multicolumn{7}{c}{XL-Sum} \\
    \cmidrule(lr){2-8} & \multicolumn{1}{c}{BLEU 3} & \multicolumn{1}{c}{BLEU 4} & \multicolumn{1}{c}{ROUGE-2} & \multicolumn{1}{c}{ROUGE-L} & \multicolumn{1}{c}{METEOR} & \multicolumn{1}{c}{mBERTScore} & \multicolumn{1}{c}{KoBERTScore} \\ 
    \midrule
    \multicolumn{1}{l}{KoGPT2$_\text{base}$} & 7.27 & 4.98 & 12.91 & 26.83 & 13.43 & 76.22 & 88.97 \\
    \multicolumn{1}{l}{KoGPT2$_\text{base}$ + \textbf{\texttt{SCRIPT}}} & \underline{7.64} & \underline{5.20} & \underline{13.30} & \underline{27.24}  & \underline{13.67} & \underline{76.31} & \underline{89.20} \\
    \midrule
    \multicolumn{1}{l}{KoGPT3-1.2B} & 9.14 & 6.21 & 15.88 & 30.18 & 16.91 & 76.95 & 89.69 \\
    \multicolumn{1}{l}{KoGPT3-1.2B + \textbf{\texttt{SCRIPT}}} & \textbf{\underline{9.39}} & \textbf{\underline{6.40}} & \textbf{\underline{15.99}} & \textbf{\underline{30.31}} & \textbf{\underline{16.97}} & \textbf{\underline{77.46}} & \textbf{\underline{89.77}} \\
    \bottomrule
  \end{tabular}
  \caption{Performance on XL-Sum multilingual summarization task. We evaluate only on the Korean summarization dataset. We use seven automatic evaluation metrics, including n-gram-based measures like BLEU, ROUGE, and METEOR; and two BERT-based scores \cite{zhang2019bertscore}, mBERTScore and KoBERTScore, for semantic similarity. The global-best results are highlighted in \textbf{boldface} and local-best results for each model are highlighted in \underline{underline}, respectively.}
  \label{tab:xl-sum}
\end{table*}

\begin{table*}[t!]
  \centering
  \setlength{\tabcolsep}{5pt}         
  \renewcommand{\arraystretch}{1.}    
  \begin{tabular}{lcccccccc}
    \toprule
    \multirow{2.5}{*}{Model} & \multicolumn{4}{c}{Kor-Learner} & \multicolumn{4}{c}{Kor-Native} \\
    \cmidrule(lr){2-5} \cmidrule(lr){6-9}
     & \multicolumn{1}{c}{$M^2_{pre}$} & \multicolumn{1}{c}{$M^2_{rec}$} & \multicolumn{1}{c}{$M^2_{F_{0.5}}$} & \multicolumn{1}{c}{GLEU} & 
    \multicolumn{1}{c}{$M^2_{pre}$} & \multicolumn{1}{c}{$M^2_{rec}$} & \multicolumn{1}{c}{$M^2_{F_{0.5}}$} & \multicolumn{1}{c}{GLEU} \\
    \midrule
    KoGPT2$_\text{base}$ & 29.35 & 16.11 & 25.19 & 21.60 & 72.12 & 55.19 & 67.76 & 61.45 \\
    KoGPT2$_\text{base}$ + \textbf{\texttt{SCRIPT}} & \underline{30.02} & \underline{16.83} & \underline{25.34} & \underline{23.54} & \underline{72.95} & \underline{56.15} & \underline{69.00} & \underline{62.25} \\
    \midrule
    KoGPT3-1.2B & 47.05 & 23.01 & 38.89 & 35.22 & 84.76 & 69.54 & 81.20 & 75.21 \\
    KoGPT3-1.2B + \textbf{\texttt{SCRIPT}} & \textbf{\underline{49.45}} & \textbf{\underline{26.36}} & \textbf{\underline{41.68}} & \textbf{\underline{39.45}} & \textbf{\underline{85.49}} & \textbf{\underline{70.02}} & \textbf{\underline{81.87}} & \textbf{\underline{75.40}} \\
    \bottomrule
  \end{tabular}
  \caption{Performance on two Korean GEC tasks. As the evaluation metrics, we use ${M^2}$ scorer \cite{dahlmeier-ng-2012-better}, which measures precision, recall, and F$_{0.5}$ scores based on edits and GLEU \cite{napoles-etal-2015-ground} for the simple n-gram matching. 
  The global-best results are highlighted in \textbf{boldface} and local-best results for each model are highlighted in \underline{underline}, respectively.}
  \label{tab:additional_korean_gec}
\end{table*}

\paragraph{Korean NLG Tasks.}
We used three distinct benchmarks, KoCommonGen, XL-Sum, and Korean Grammatical Error Correction (GEC), to evaluate the performance of our proposed method on Korean NLG tasks. The detailed explanations of each benchmark are provided below: 
\begin{itemize}[leftmargin=15px]
    \item KoCommonGen \cite{seo-etal-2022-dog}: A generative commonsense reasoning dataset comprising 43,188 train samples and 2,040 test examples. Given a set of morphemes, the model composes a sentence that reflects commonsense knowledge.
    \item XL-Sum \cite{hasan-etal-2021-xl}: A summarization dataset used to evaluate models' ability to generate concise and accurate summaries from large text bodies. We focus on the Korean subset, which includes 4,407 train samples and 550 validation and test samples. We are only using the Korean subset of XL-Sum dataset for our experiments.
    \item Korean GEC \cite{yoon-etal-2023-towards}: A grammatical error correction dataset for Korean. It consists of four sub-datasets: three standalone datasets, such as Kor-Learner, Kor-Lang8, and Kor-Native, and one aggregated dataset, Kor-Union. Kor-Learner offers a more structured and reliable dataset for the Korean GEC task, as it is annotated by Korean language tutors. In contrast, Kor-Lang8 was corrected by native speakers through an open online platform. In this paper, we focus on two orthogonal GEC tasks, such as Kor-Learner and Kor-Native, to more accurately evaluate model performance relative to dataset complexity.
    \begin{itemize} [leftmargin=10px]
        \item Kor-Learner GEC \cite{yoon-etal-2023-towards}: A GEC dataset for Korean learner texts, containing 19,898 train sentences, 4,264 validation sentences, and 4,265 test sentences. Kor-Learner contains learner-written essays that have been carefully corrected and annotated by Korean tutors. It aids in identifying and correcting grammar errors specific to Korean language learners.
        \item Kor-Native GEC \cite{yoon-etal-2023-towards}: A GEC dataset targeting native Korean texts to support advanced linguistic understanding. It comprises 12,292 train sentences, 2,634 validation sentences, and 2,634 test sentences.
    \end{itemize}
\end{itemize}

\section{Evaluation on Generative Tasks}
\label{app:extra_gen_tasks}
\subsection{XL-Sum}
As shown in Table~\ref{tab:xl-sum}, \textbf{\texttt{SCRIPT}} consistently demonstrated its effectiveness on n-gram metrics. However, the performance improvement observed in the summarization task was somewhat smaller compared to that in the commonsense generation task. This difference arises because the summarization task typically involves input and output sentences that are approximately five times longer than those in the commonsense generation task, such as KoCommonGen. This suggests that \textbf{\texttt{SCRIPT}} is particularly effective at generating concise, well-formed sentences, demonstrating its strength in handling shorter and more focused outputs.

\begin{table*}[t!]
  \centering
  \renewcommand{\arraystretch}{1.}    
  \begin{tabular}{lcccccc}
    \toprule
    \multirow{2.5}{*}{Model} & \multicolumn{5}{c}{KoBEST} & \multirow{2.5}{*}{Avg.} \\
    \cmidrule(lr){2-6}
    & \multicolumn{1}{c}{BoolQ} & \multicolumn{1}{c}{COPA} & \multicolumn{1}{c}{WiC} & \multicolumn{1}{c}{HellaSwag} & \multicolumn{1}{c}{SentiNeg} &  \\
    \midrule
    \multicolumn{1}{l}{KoGPT3-1.2B} & 77.32 & \underline{\textbf{82.80}} & 72.78 & 78.90 & 96.31 & 81.62 \\
    \multicolumn{1}{l}{KoGPT3-1.2B + \textbf{\texttt{SCRIPT}}} & \underline{\textbf{77.63}} & \underline{\textbf{82.80}} & \underline{\textbf{74.65}} & \underline{\textbf{79.30}} & \underline{\textbf{96.48}} & \underline{\textbf{82.17}} \\
    \midrule
    \multicolumn{1}{l}{mGPT-1.3B} & \underline{71.19} & 69.70 & 68.38 & 76.10 & 89.40 & 74.95 \\
    \multicolumn{1}{l}{mGPT-1.3B + \textbf{\texttt{SCRIPT}}} & 70.72 & \underline{70.60} & \underline{69.17} & \underline{76.70} & \underline{90.42} & \underline{75.52} \\
    \bottomrule
  \end{tabular}
  \caption{Performance of KoGPT3-1.2B and mGPT-1.3B on KoBEST benchmark. The evaluation metrics for each task are accuracy (\%). The global-best results are highlighted in \textbf{boldface} and local-best results for each model are highlighted in \underline{underline}, respectively.}
  \label{mgpt_1}
\end{table*}

\begin{table*}[t!]
  \centering
  \footnotesize
  \setlength{\tabcolsep}{2pt}         
  \renewcommand{\arraystretch}{1.3}    
  \begin{tabular}{lcccccccc}
    \toprule
    \multirow{2.5}{*}{Model} & \multicolumn{7}{c}{KoCommonGen} & \multirow{2.5}{*}{Avg.} \\
    \cmidrule(lr){2-8} & 
    \multicolumn{1}{c}{BLEU 3} & \multicolumn{1}{c}{BLEU 4} & \multicolumn{1}{c}{ROUGE-2} & \multicolumn{1}{c}{ROUGE-L} & \multicolumn{1}{c}{METEOR} & \multicolumn{1}{c}{mBERTScore} & \multicolumn{1}{c}{KoBERTScore}\\
    \midrule
    \multicolumn{1}{l}{KoGPT3-1.2B} & 26.19 & 17.20 & 58.85 & 62.53 & 52.11 & 85.41 & 91.17 & 56.21 \\
    \multicolumn{1}{l}{KoGPT3-1.2B + \textbf{\texttt{SCRIPT}}} & \textbf{\underline{28.89}} & \textbf{\underline{19.58}} & \textbf{\underline{59.28}} & \textbf{\underline{64.80}} & \textbf{\underline{52.37}} & \textbf{\underline{86.26}} & \textbf{\underline{91.78}} & \textbf{\underline{57.57}} \\
    \midrule
    \multicolumn{1}{l}{mGPT-1.3B} & 15.16 & 8.07 & 37.77 & 50.99 & 33.31 & 80.17 & 89.34 & 44.97 \\
    \multicolumn{1}{l}{mGPT-1.3B + \textbf{\texttt{SCRIPT}}} & \underline{16.59} & \underline{9.11} & \underline{39.31} & \underline{52.68} & \underline{34.77} & \underline{80.82} & \underline{89.95} & \underline{46.18} \\
    \bottomrule
  \end{tabular}
  \caption{Performance of KoGPT3-1.2B and mGPT-1.3B on KoCommonGen dataset. We use eight automatic evaluation metrics: BLEU, ROUGE, and METEOR for n-gram-based measures; and mBERTScore and KoBERTScore for semantic similarity. The global-best results are highlighted in \textbf{boldface} and local-best results for each model are highlighted in \underline{underline}, respectively.}
  \label{mgpt_2}
\end{table*}

\subsection{Korean GEC}
Our proposed method, \textbf{\texttt{SCRIPT}}, also demonstrated the strongest performance on Korean grammatical error correction tasks. As shown in Table~\ref{tab:additional_korean_gec}, it consistently outperformed the base model in both the Kor-Learner and Kor-Native tasks, showing particularly strong effectiveness in the Kor-Learner task with average improvements exceeding an average of 3.2\%p gains over the global-best performing baseline.
According to \citet{yoon-etal-2023-towards}, the Kor-Learner task contains a large proportion of errors related to particles, endings, and conjugations compared to the Kor-Native task. As mentioned in Section~\ref{sec:motivation}, linguistic variations in Korean frequently occur at the subcharacter-level. Therefore, the substantial performance gains observed on the Kor-Learner task demonstrate the effectiveness of our core approach: integrating subcharacter compositional information into subword representations. This result further validates that \textbf{\texttt{SCRIPT}} is highly suitable for the Korean language and effectively enriches the PLM's subword representations.

\section{Effectiveness of Multilingual Model}
\label{app:multilingual}
Our proposed method, \textbf{\texttt{SCRIPT}}, can be seamlessly integrated into the embedding layer of any model and is broadly applicable in multilingual settings. To demonstrate its effectiveness for Korean in multilingual models, we evaluated it on both a Korean monolingual model and a multilingual model that supports Korean. Specifically, we used KoGPT3-1.2B as the monolingual baseline and mGPT-1.3B, a multilingual model with comparable architecture and scale. For the NLU task, we employed the KoBEST benchmark to assess knowledge understanding, and for the NLG task, we used KoCommonGen to evaluate complex knowledge generation, including commonsense reasoning.

As shown in Table~\ref{mgpt_1} and Table~\ref{mgpt_2}, multilingual models with \textbf{\texttt{SCRIPT}} largely outperformed their base counterparts in both KoBEST (NLU) and KoCommonGen (NLG) tasks. 
Notably, mGPT-1.3B achieved a performance gain of approximately 0.6\%p on KoBEST benchmarks and 1.2\%p on KoCommonGen, closely mirroring the improvements observed in the monolingual model. Gains in generative tasks were nearly twice as large as those in understanding tasks, indicating that \textbf{\texttt{SCRIPT}} is particularly effective in enhancing generative capabilities for Korean. These findings highlight the promise of scaling up to significantly larger and more extensively pre-trained multilingual generative models, such as the Llama~\cite{dubey2024llama3herdmodels} and Qwen~\cite{qwen2.5_2025} series.
In particular, this substantial improvement in multilingual models is especially valuable given the current scarcity of specialized pre-trained LLMs for Korean.

\section{Qualitative Analysis for Generations}
\label{app:qualitative}
To analyze the quality of machine-generated text, in Table~\ref{tab:qualitative_1}, we conducted a qualitative analysis for each generative task by model. We established two baselines for comparison: the basic KoGPT2$_\text{base}$ model and KoGPT2$_\text{base}$+\textbf{\texttt{SCRIPT}}. This analysis covered all Korean NLG tasks performed in Section~\ref{sec:korean_nlg_tasks}. Since the input text for the summarization task, XL-Sum, is quite long, we have included examples in the \ref{app:qual_for_xl_sum} for further details if needed.

\subsection{KoCommonGen}
\label{app:qual_for_kocommongen}
Given the set of morphemes as input to the model, it generates the sentence as output by including the morphemes. As a result of the experiments, shown in Table~\ref{tab:qualitative_1}, we observed that KoGPT2$_\text{base}$+\textbf{\texttt{SCRIPT}} model correctly generated appropriate particles by identifying the characteristics and position of objects, such as rail, train, and road. In contrast, the baseline model, KoGPT2$_\text{base}$, incorrectly generated the position of the word `train' as `beside the tracks' instead of `on the tracks'. In English, the difference between 'beside' and 'on' involves several letters, whereas in Korean, this distinction is very subtle, differing by only a single subcharacter, `ㅢ' (`옆의') for `on' and `ㅔ' (`옆에') for `beside', which makes it more challenging to distinguish. This shows that \textbf{\texttt{SCRIPT}} effectively captures subtle nuances at the subcharacter-level.

\subsection{XL-Sum}
\label{app:qual_for_xl_sum}
The summarization performance appears similar, but the base model tends to generate slightly longer sentences. Overall, our proposed method produces more concise summaries. For example, in this task’s sample data, while KoGPT2$_\text{base}$ focused on the 'act of collecting samples', including the 'lunar landing', our model emphasized the 'successful completion of the exploration', generating sentences with a clearer focus on summarization itself.

\subsection{Kor-Learner}
\label{app:qual_for_kor-learner}
As mentioned earlier in Section~\ref{sec:korean_nlg_tasks}, the Kor-Learner dataset contains a higher frequency of errors related to particles, endings, and conjugations compared to other Korean GEC datasets. The sample in Table~\ref{tab:qualitative_1} also requires corrections for grammatical errors in endings. While the KoGPT2$_\text{base}$ model failed to correct these properly, our proposed method successfully adjusted endings by considering their agreement with predicates. As shown in Figure~\ref{fig:Korean_Example}, understanding these types of grammatical errors is particularly important for Korean. Thus, our proposed method is both suitable and essential for effectively handling Korean.

\subsection{Kor-Native}
\label{app:qual_for_kor-native}
Through examples from the Kor-Native task, we confirmed that our proposed method, \textbf{\texttt{SCRIPT}}, enhances the ability to handle whitespace and noun recognition effectively. As shown in the sample in Table~\ref{tab:qualitative_1}, it asked to identify and correct the incorrect word `테니그$_{tennig}$' to the appropriate noun `테니스$_{tennis}$'. While the naive KoGPT2$_\text{base}$ model failed to detect the error in this sentence and thus cannot make the necessary correction, the model using our proposed \textbf{\texttt{SCRIPT}} method successfully identified and corrected wrong word to `테니스$_{tennis}$'. Although this adjustment involved only a subtle subcharacter-level difference, changing `ㄱ$_{g}$' to `ㅅ$_{s}$', it once again demonstrated that subword models using larger token units than character-level cannot adequately handle such distinctions.

\section{Impact of Fused Representations}
\label{app:integrated_embeddings}
\begin{figure*}[t!]
  \centering
  \includegraphics[width=0.96\linewidth]{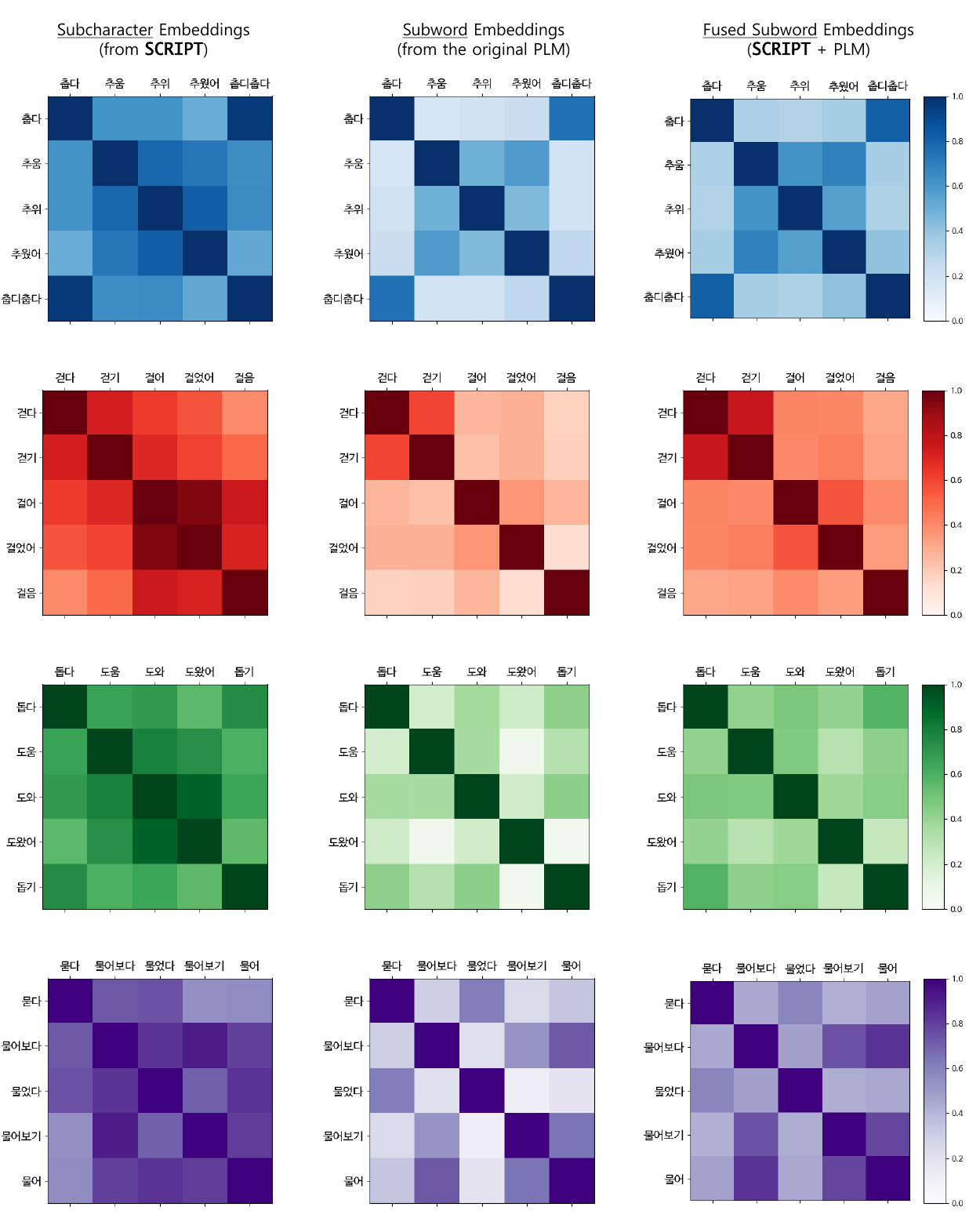}
  \caption{Visualization of the similarities between word representations, measured from conjugated word pairs. Each word set contains five words that share the same root meaning. The first word set, \{`춥다', `추움', `추위', `추웠어', `춥디춥다'\}, conveys the meaning `cold'; the second set, \{`걷다', `걷기', `걸어', `걸었어', `걸음'\}, represents `walk'; the third set, \{`돕다', `도움', `도와', `도왔어', `돕기'\}, signifies `help'; and the final set, \{`묻다', `물어보다', `물었다', `물어보기', `물어'\}, conveys the meaning `ask'. Using these word sets, we compared three different types of embeddings: those derived from subcharacter embeddings in \textbf{\texttt{SCRIPT}}, subword embeddings from the PLM, and fusion embeddings augmented by \textbf{\texttt{SCRIPT}}.}
  \label{fig:subword_embeddings}
\end{figure*}

To examine how compositional knowledge of subcharacters affects subword representations, we compare (i) subcharacter embeddings from \textbf{\texttt{SCRIPT}}, (ii) original subword embeddings from the PLM, and (iii) fused subword embeddings augmented by \textbf{\texttt{SCRIPT}}. 
As shown in Figure~\ref{fig:subword_embeddings}, \textbf{\texttt{SCRIPT}}’s subcharacter representations yield the highest similarity among the predicate inflected word sets sharing root semantics but differing at the subcharacter-level. Notably, this advantage transfers to the fused embeddings, which accurately capture these fine-grained relational patterns. These results underscore the effectiveness of the proposed module in modeling subcharacter-level variation through compositional and representational fusion.

\clearpage

\begin{table*}[t!]
\small
\centering
\renewcommand{\arraystretch}{0.5}
\begin{tabular}{|c|c|c|}
  \hline
  \multirow{2.5}{*}{\textbf{Task}} & \multirow{2.5}{*}{\textbf{Lang}} & \multirow{2.5}{*}{\textbf{Example}} \\ 
   & & \\
  \hline
  \multirow{12}{*}{KoCommonGen} & Ko & \begin{tabular}{p{11.5cm}}
    \begin{description}
      \item[Input:] \{ 있, 선로, 길, 옆, 열차 \}
      \item[Gold Label:] 길 \textbf{\textcolor{blue}{옆의}} 선로\textbf{\textcolor{blue}{에}} 열차가 있다.
      \item[KoGPT2:] 선로가 길 \textbf{\textcolor{blue}{옆의}} 길 \textbf{\textcolor{red}{옆에}} 열차가 세워져 있다.
      \item[KoGPT2 + \textbf{\texttt{SCRIPT}}:] 열차들이 길 \textbf{\textcolor{blue}{옆의}} 선로\textbf{\textcolor{blue}{에}} 있다.
    \end{description}
    \end{tabular} \\
    \cline{2-3}
    & En & \begin{tabular}{p{11.5cm}}
    \begin{description}
      \item[Input:] \{ be, tracks, road, beside, train \}
      \item[Gold Label:] There is a train \textbf{\textcolor{blue}{on}} the tracks \textbf{\textcolor{blue}{beside}} the road.
      \item[KoGPT2:] The train is parked \textbf{\textcolor{red}{beside}} the track \textbf{\textcolor{blue}{beside}} the road.
      \item[KoGPT2 + \textbf{\texttt{SCRIPT}}:] The trains are \textbf{\textcolor{blue}{on}} the tracks \textbf{\textcolor{blue}{beside}} the road.
    \end{description}
    \end{tabular} \\
  \hline
  \multirow{20}{*}{Kor-Learner} & Ko & \begin{tabular}{p{11.5cm}}
    \begin{description}
      \item[Input:] 그리고 가장 중요한 영향은 그 앞으로 그 여행으로 이전보다 훨씬 더 `처음' 을 접할 \textbf{\textcolor{red}{거다}}.
      \item[Gold Label:] 그리고 가장 중요한 영향은 앞으로 여행으로 이전보다 훨씬 더 `처음' 을 접할 \textbf{\textcolor{blue}{것이라는 사실이다}}.
      \item[KoGPT2:] 그리고 가장 중요한 영향을 그 앞으로 그 여행으로 이전보다 훨씬 더 `처음' 을 접할 \textbf{\textcolor{red}{거다}}.
      \item[KoGPT2 + \textbf{\texttt{SCRIPT}}:] 그리고 가장 중요한 영향은 그 앞으로 그 여행으로 이전보다 훨씬 더 `처음' 을 접할 \textbf{\textcolor{blue}{거라는 것이다}}.
    \end{description}
    \end{tabular} \\
    \cline{2-3} 
    & En & \begin{tabular}{p{11.5cm}}
    \begin{description}
      \item[Input:] And the most important impact \textbf{\textcolor{red}{would}} that, through that journey, they will encounter the `first' much more than before.
      \item[Gold Label:] And the most important impact \textbf{\textcolor{blue}{is the fact}} that, through future travels, they will encounter the `first' much more than before.
      \item[KoGPT2:] And the most important impact \textbf{\textcolor{red}{would}} that, through that journey, they will encounter the `first' much more than before.
      \item[KoGPT2 + \textbf{\texttt{SCRIPT}}:] And the most important impact \textbf{\textcolor{blue}{is the thing}} that, through that journey, they will encounter the `first' much more than before.
    \end{description}
    \end{tabular} \\
  \hline
  \multirow{12}{*}{Kor-Native} & Ko & \begin{tabular}{p{11.5cm}}
    \begin{description}
      \item[Input:] 주말에 함께 \textbf{\textcolor{red}{*테니그}}를 쳐여.
      \item[Gold Label:] 주말에 함께 \textbf{\textcolor{blue}{테니스}}를 쳐요.
      \item[KoGPT2:] 주말에 함께 \textbf{\textcolor{red}{*테니그}}를 쳐요.
      \item[KoGPT2 + \textbf{\texttt{SCRIPT}}:] 주말에 함께 \textbf{\textcolor{blue}{테니스}}를 쳐요.
    \end{description}
    \end{tabular} \\
    \cline{2-3}
    & En & \begin{tabular}{p{11.5cm}}
    \begin{description}
      \item[Input:] Let's play \textbf{\textcolor{red}{*tennig}} together on the weekend.  
      \item[Gold Label:] Let's play \textbf{\textcolor{blue}{tennis}} together on the weekend.
      \item[KoGPT2:] Let's play \textbf{\textcolor{red}{*tennig}} together on the weekend.
      \item[KoGPT2 + \textbf{\texttt{SCRIPT}}:] Let's play \textbf{\textcolor{blue}{tennis}} together on the weekend.
    \end{description}
    \end{tabular} \\
  \hline
  \end{tabular}
  \caption{Examples for the qualitative analysis of four generation tasks. For each task, examples are composed of the input provided to the model, the gold label, and the predictions generated by two baselines: KoGPT2$_\text{base}$ and KoGPT2$_\text{base}$ applied with \textbf{\texttt{SCRIPT}}. The model outputs were generated in Korean, with English translations provided alongside for clarity. The asterisk (*) indicates a ungrammatical word. \textcolor{red}{Red-colored} characters represent incorrect parts, while \textcolor{blue}{blue-colored} characters indicate correct representations.}
  \label{tab:qualitative_1}
\end{table*}

\begin{table*}[t!]
\small
\centering
\renewcommand{\arraystretch}{1.}
\begin{tabular}{|c|c|c|}
  \hline
  \multirow{2}{*}{\textbf{Task}} & \multirow{2}{*}{\textbf{Language}} & \multirow{2}{*}{\textbf{Example}} \\ 
   & & \\
  \hline
  \multirow{23}{*}{XL-Sum} & Korean & \begin{tabular}{p{11.5cm}}
    \begin{description}
      \item[Input:] 최종 목적은 2㎏ 정도의 `토양' 표본을 상승선, 귀환선에 전달해 지구까지 가져오는 것이다 중국국가우주국(CNSA)은 달의 암석과 토양 표본을 수집해 지구로 가져오기 위해 출발한 무인 달 탐사선 `창어 5호'가 1일 밤 착륙에 성공했다고 2일 밝혔다. 창어 5호는 `폭풍의 바다'(Oceanus Procellarum)라는 지역 내 `몽스 륌케르'(Mons Rümker) 화산지대 북쪽에 안착했다. 이곳에서 며칠간 달 표면의 흙과 암석 표본 등을 수집한다. 창어 5호 탐사선에는 작업을 돕기 위한 카메라, 레이더, 드릴, 삽 등이 탑재돼있다. 최종 목적은 2㎏ 정도의 표토 표본을 상승선과 궤도선을 거쳐 귀환선에 전달해 지구까지 가져오는 것이다. 달의 토양 표본을 지구로 가져온 탐사선은 44년 전 1976년 옛 소련의 루나 24호가 마지막으로, 당시 200g의 토양을 지구로 옮기는 데 성공했다. 창어 5호 프로젝트 팀이 환호하는 모습 이날 달 착륙 모습은 일주일 전 발사 때와 달리 생중계 되지 않았다. 중국 TV 채널에서는 성공적인 착륙이 확인되고 나서야 정규 방송을 중단하고 이를 녹화 중계했다. 공개된 착륙 과정에는 탐사선의 다리가 달의 먼지 쌓인 표면에 그림자를 드리우는 장면 등이 포함됐다. ...
      \item[KoGPT2:] 중국 국가우주국이 달 착륙에 성공한 창어 6호 달 착륙선에 탑재된 카메라와 레이더를 통해 달 표면 표본을 지구까지 운반했다.
      \item[KoGPT2 + \textbf{\texttt{SCRIPT}}:] 중국의 달 탐사 프로젝트가 성공적으로 마무리됐다.
    \end{description}
    \end{tabular} \\
    \cline{2-3}
    & English & \begin{tabular}{p{11.5cm}}
    \begin{description}
      \item[Input:] The primary goal is to bring approximately 2 kg of lunar soil samples back to Earth by transferring them from the ascent and return modules. The China National Space Administration (CNSA) announced on the 2nd that its unmanned lunar probe, Chang’e-5, successfully landed on the night of the 1st to collect lunar rock and soil samples to return to Earth. Chang’e-5 has landed in the volcanic area north of Mons Rümker within the region known as Oceanus Procellarum. Over the next few days, it will collect samples of lunar soil and rock. Equipped with cameras, radar, drills, and shovels to aid in its operations, the ultimate goal of Chang’e-5 is to gather about 2 kg of surface samples, which will be transferred from the ascent and orbital modules to the return module for their journey back to Earth. The last mission to bring lunar soil samples to Earth was the Soviet Union's Luna 24 in 1976, which successfully transported 200 g of lunar soil back to Earth. On the day of the landing, the Chang’e-5 team celebrated. Unlike the launch, the landing was not broadcast live, and Chinese TV channels interrupted regular programming to air recorded footage after confirming a successful landing. The released landing process included images of the lander casting a shadow on the dusty lunar surface. ...
      \item[Gold Label:] China has landed another probe on the surface of the moon.
      \item[KoGPT2:] The China National Space Administration successfully transported lunar surface samples to Earth using the camera and radar aboard the Chang'e 6 lunar lander.
      \item[KoGPT2 + \textbf{\texttt{SCRIPT}}:] China's lunar exploration project has been successfully completed.
    \end{description}
    \end{tabular} \\
    \hline
  \end{tabular}
  \caption{Examples for the qualitative analysis of XL-Sum task. For each task, examples are composed of the input provided to the model, the gold label, and the predictions generated by two baselines: KoGPT2$_\text{base}$ and KoGPT2$_\text{base}$ applied with \textbf{\texttt{SCRIPT}}. The model outputs were generated in Korean, with English translations provided alongside for clarity.}
  \label{tab:qualitative_2}
\end{table*}

\clearpage
\section{Computational Efficiency}
\label{app:computational_efficiency}
\subsection{Computational Complexity}
\label{app:computational_complexity}

\begin{table*}[h!]
  \centering
  \setlength{\tabcolsep}{5pt}         
  \renewcommand{\arraystretch}{1.}    
  \begin{tabular}{lccc}
    \toprule
    Model & Embedding Layer & Transformer Stacks & Restoration Layer \\
    \midrule
    BERT & $O(1)$ & $O({N_\text{s}}^2D)$ & -\\
    BERT + \textbf{\texttt{SCRIPT}} & $O(N_\text{j}D^2 + {N_\text{s}}^2D$) & $O({N_\text{s}}^2D)$ & -\\
    KOMBO & $O(N_\text{j}D^2 + {N_\text{j}}^2D)$ & $O({N_\text{c}}^2D)$ & $O(N_\text{c}D^2)$ \\
    \bottomrule
  \end{tabular}
  \caption{Comparison of the computational complexities across the three components of the model's architecture. $N_\text{j}$ is the length of subcharacter sequence, $N_\text{s}$ is the length of subword sequence, $N_\text{c}$ is the length of character sequence, and $D$ is the hidden size.}
  \label{tab:complexity}
\end{table*}

As shown in Table~\ref{tab:complexity}, we quantify the overhead of each approach by breaking computation into three components: the embedding layer, the Transformer stack, and any model-specific layers.

In the embedding layer, standard BERT performs a simple embedding lookup and linear projection, which has constant time complexity with respect to sequence length. Adding \textbf{\texttt{SCRIPT}} introduces modest overhead: it applies a GRU-based encoder to contextualize the subcharacter sequence for each token and uses cross-attention to fuse this information with the original subword embedding. Both components are single-layer operations, in contrast to the deep Transformer stack with more than 10 layers. Moreover, \textbf{\texttt{SCRIPT}} compresses the subcharacter sequence before cross-attention, keeping sequence lengths short during the expensive fusion step. By comparison, KOMBO’s embedding stage is significantly heavier (N$\text{s}$ ≪ N$\text{j}$). It processes the full subcharacter sequence with a GRU, three self-attention layers, and another GRU for compression, making its embedding computation far more costly than \textbf{\texttt{SCRIPT}}. In short, both methods add overhead beyond the base model, but \textbf{\texttt{SCRIPT}} is much lighter due to its efficient compression and fusion strategy.

In the Transformer stack, \textbf{\texttt{SCRIPT}} again aligns closely with the base model. Both the base model (BERT) and BERT+\textbf{\texttt{SCRIPT}} operate at the subword level throughout the Transformer layers. This means their self-attention complexity scales with the subword sequence length ($O({N_\text{s}}^2 D)$ per layer, where N$_\text{s}$ is the number of subword tokens and D is the hidden dimension), just as in the original model. In contrast, KOMBO converts inputs into much longer character-level sequences (N$\text{c}$ ≫ N$\text{s}$), yielding a per-layer complexity of $O({N\text{c}}^{2}D)$. This makes KOMBO’s Transformer blocks slower and more memory-intensive for the same input. 

Additionally, KOMBO requires a restoration layer after the Transformer stack to convert character-level outputs back to subword representations, implemented with another GRU. Neither BERT nor BERT+\textbf{\texttt{SCRIPT}} requires such steps. Thus, aside from a small embedding-stage overhead, BERT+\textbf{\texttt{SCRIPT}} preserves the base model’s computational profile, whereas KOMBO incurs substantial extra cost in both the Transformer and output stages.

\subsection{Computational Cost}
\label{app:computational_cost}
To assess how the computational complexity discussed in the previous section translates into actual computing cost, we empirically evaluate the architectural differences in terms of GPU memory usage and training time.

\label{computational_complexity}
\begin{figure}[h]
  \centering
  \subfigure[]{
    \includegraphics[width=1.\linewidth]{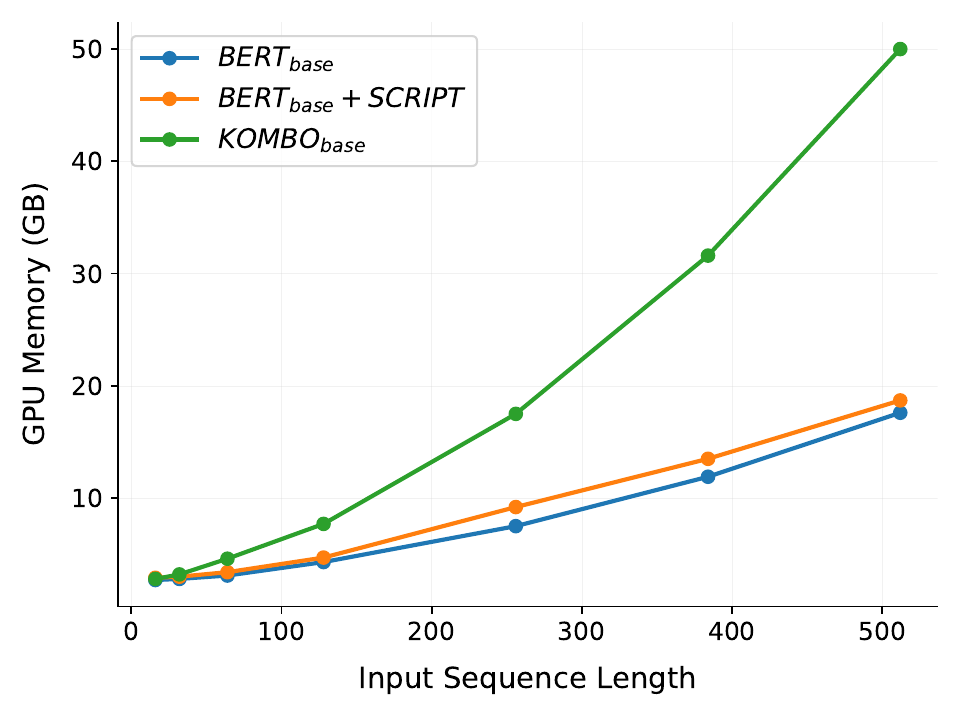}
    \label{fig:gpu_memory}
  }
  \subfigure[]{
    \includegraphics[width=1.\linewidth]{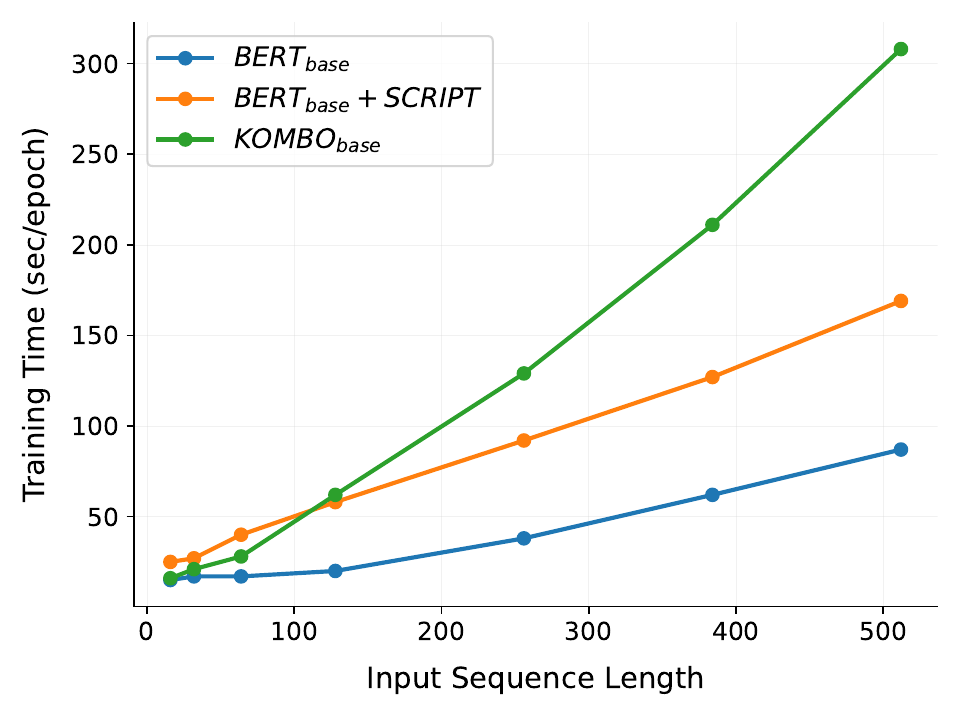}
    \label{fig:training_time}
  }
  \caption{
  Comparison of computational costs among the base model BERT$_\text{base}$, BERT$_\text{base}$ with \textbf{\texttt{SCRIPT}}, and previous Jamo-based PLM, KOMBO$_\text{base}^\text{Jamo}$. The results were obtained on the KB-HellaSwag benchmark, a representative Korean NLU task, using a single NVIDIA RTX 3090 GPU. (a) Peak GPU memory usage during training with varying input sequence lengths. (b) Training time per epoch with varying input sequence lengths.
  }
  \label{fig:computational_cost}
\end{figure}

\paragraph{GPU Memory.}
Figure~\ref{fig:gpu_memory} summarizes the resource footprint of each model across varying input sequence lengths. The GPU memory consumption of BERT$_\text{base}$+\textbf{\texttt{SCRIPT}} is nearly identical to that of BERT$_\text{base}$ alone, and it remains far lower than that of KOMBO$_\text{base}$, especially for longer sequences. For instance, at an input length of 256 tokens, the BERT$_\text{base}$ model uses about 7.5 GB of GPU memory during training, and BERT$_\text{base}$+\textbf{\texttt{SCRIPT}} requires approximately 9.2 GB, a relatively small 1.7 GB increase. In contrast, KOMBO$_\text{base}$ at the same sequence length demands roughly 17.5 GB - more than double the memory of the base model. This gap widens with longer inputs: at 512 tokens, BERT$_\text{base}$+\textbf{\texttt{SCRIPT}} uses 18.7 GB vs. 17.6 GB for BERT (only a 6\% increase), whereas KOMBO$_\text{base}$ soars to about 50 GB, nearly three times the base model’s requirement. These results confirm that plug-in design of \textbf{\texttt{SCRIPT}} adds minimal memory overhead, while KOMBO’s character-level processing and extra layers drastically inflate memory usage for large inputs.

\paragraph{Training Time per Epoch.}
A similar pattern is observed in training time. As shown in Figure~\ref{fig:training_time}, \textbf{\texttt{SCRIPT}} introduces only a moderate slowdown relative to the base model, whereas KOMBO$_\text{base}$ dramatically reduces training speed as sequence length grows. For a moderate input length (128 tokens), BERT$_\text{base}$+\textbf{\texttt{SCRIPT}} requires roughly 58 s per training step compared to 20 s for BERT$_\text{base}$ (about 2.9× slower), and KOMBO$_\text{base}$ takes around 62 s (about 3.1× slower than base). However, as the sequence length increases, KOMBO’s runtime cost grows much more rapidly. At 512 tokens, BERT$_\text{base}$+\textbf{\texttt{SCRIPT}} processes a batch in roughly 169 s (less than 2× the ~87 s required by BERT$_\text{base}$), whereas KOMBO$_\text{base}$ requires about 308 s – over 3.5× the base model’s time. This steep slowdown for KOMBO is a direct consequence of operating over a much longer sequence with additional transformation layers, as discussed above. In contrast, \textbf{\texttt{SCRIPT}} maintains a moderate runtime overhead, less than 2× the base model even at maximum sequence lengths, making it far more practical than KOMBO in real-world training scenarios.

We note that these trends hold for both encoder-based and decoder-based architectures. In our experiments with the decoder-only KoGPT2 model, adding \textbf{\texttt{SCRIPT}} incurred similar slowdowns and only minor memory increases, underscoring the general applicability of \textbf{\texttt{SCRIPT}} across model types.

In summary, \textbf{\texttt{SCRIPT}} offers a significantly more efficient and practical solution for incorporating subcharacter information than another subcharacter-based approach, such as KOMBO. By sidestepping expensive architectural changes and pre-training requirements, \textbf{\texttt{SCRIPT}} maintains almost the same training footprint as the underlying base model in terms of memory. The small overhead introduced by \textbf{\texttt{SCRIPT}} is significantly outweighed by its benefits, and it stands in stark contrast to the heavy computational cost of KOMBO. This efficiency makes \textbf{\texttt{SCRIPT}} a highly practical plug-and-play module for real-world deployment on large-scale models and datasets. Next, we examine another aspect of training efficiency, the convergence speed of each model during training, to further assess the practical advantages of \textbf{\texttt{SCRIPT}}.

\begin{figure*}[h!]
    \centering
    \includegraphics[width=1.\linewidth]{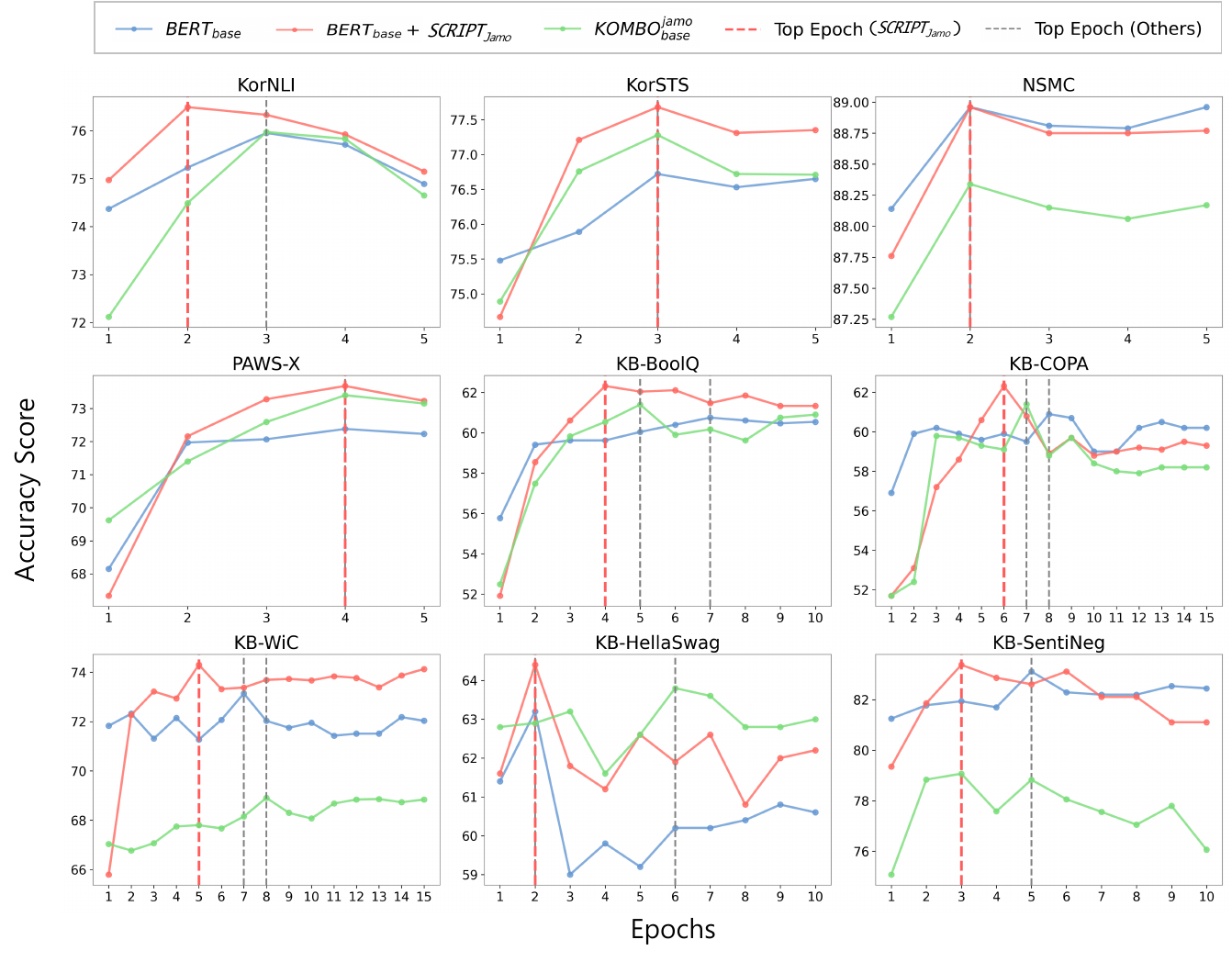}
    \caption{
    The graphs show the fine-tuning performances of three models, BERT$_\text{base}$, BERT$_\text{base}$+\textbf{\texttt{SCRIPT}}, and KOMBO$_\text{base}$, across nine NLU tasks. The x-axis represents the number of epochs for each task, and the y-axis indicates the accuracy score. Two types of dotted vertical lines are overlaid on the line graphs for each model: the red dotted line marks the epoch at which the model with the proposed \textbf{\texttt{SCRIPT}} module achieves its best performance, while the gray dotted lines indicate the best-performing epochs for the other two models.
    }
    \label{fig:nlu_per_eps}
\end{figure*}

\subsection{Training Efficiency}
\label{app:training_efficiency}

In Section~\ref{computational_complexity}, we analyzed the structural efficiency of the proposed method \textbf{\texttt{SCRIPT}} compared to another off-the-shelf subcharacter-based model, KOMBO. 
In this section, we further investigate training efficiency, focusing on the convergence speed across three different models, such as BERT$_\text{base}$ as the base model, BERT$_\text{base}$ + \textbf{\texttt{SCRIPT}}, and KOMBO$_\text{base}$, during fine-tuning. Figure~\ref{fig:nlu_per_eps} compares the convergence points of each model across the nine principal Korean NLU tasks introduced in Section~\ref{sec:korean_standard_nlu_tasks}. As a result, BERT$_\text{base}$+\textbf{\texttt{SCRIPT}} consistently converges faster as both BERT$_\text{base}$ and KOMBO$_\text{base}$, achieving superior performance with fewer training epochs. Specifically, it converged faster on six out of nine tasks, and matched the convergence speed on the remaining three tasks.
These results demonstrate that our method significantly enhances learning efficiency. Furthermore, though \textbf{\texttt{SCRIPT}} leverages both subword and subcharacter embeddings, it not only outperforms the model utilizing solely 32k subwords (BERT$_\text{base}$) and fewer than 200 subcharacters (KOMBO$_\text{base}$) but also converges more rapidly, highlighting its strong adaptation to Korean language understanding.

Overall, our comprehensive evaluation demonstrates that \textbf{\texttt{SCRIPT}} offers a substantially more efficient and scalable alternative to prior subcharacter-based approaches. By injecting subcharacter compositional knowledge directly into existing PLM embeddings, \textbf{\texttt{SCRIPT}} enriches the model’s representational capacity while preserving the computational profile of the base model. This dual advantage, greater linguistic expressiveness with only marginal computational overhead, establishes \textbf{\texttt{SCRIPT}} as a practical and robust solution for real-world deployment in Korean NLP.

\end{CJK}

\end{document}